\documentclass{IEEEtran}
\usepackage[utf8]{inputenc}

\usepackage{hyperref}

\usepackage{pgfplots}
\pgfplotsset{compat=1.18}

\usepackage{amssymb}
\usepackage{amsmath}

\usepackage{subfigure} % subcaptions for subfigures
\usepackage{subfigmat} % matrices of similar subfigures

\usepackage{verbatim}

\usepackage{tikz}

\usepackage{xcolor}
\definecolor{fc}{HTML}{1E90FF}
\definecolor{h}{HTML}{228B22}
\definecolor{bias}{HTML}{87CEFA}
\definecolor{noise}{HTML}{8B008B}
\definecolor{conv}{HTML}{FFA500}
\definecolor{pool}{HTML}{B22222}
\definecolor{up}{HTML}{B22222}
\definecolor{view}{HTML}{FFFFFF}
\definecolor{bn}{HTML}{FFD700}
\tikzset{fc/.style={black,draw=black,fill=fc,rectangle,minimum height=1cm}}
\tikzset{h/.style={black,draw=black,fill=h,rectangle,minimum height=1cm}}
\tikzset{bias/.style={black,draw=black,fill=bias,rectangle,minimum height=1cm}}
\tikzset{noise/.style={black,draw=black,fill=noise,rectangle,minimum height=1cm}}
\tikzset{conv/.style={black,draw=black,fill=conv,rectangle,minimum height=1cm}}
\tikzset{pool/.style={black,draw=black,fill=pool,rectangle,minimum height=1cm}}
\tikzset{up/.style={black,draw=black,fill=up,rectangle,minimum height=1cm}}
\tikzset{view/.style={black,draw=black,fill=view,rectangle,minimum height=1cm}}
\tikzset{bn/.style={black,draw=black,fill=bn,rectangle,minimum height=1cm}}

\title{Deep-seeded Clustering for Emotion Recognition from Wearable Physiological Sensors}
\author{Marta A. Conceição, Antoine Dubois, Sonja Haustein, Bruno Miranda, Carlos Lima Azevedo}
\date{April 2025}

\begin{document}

\maketitle
\begin{abstract}
According to the circumplex model of affect, an emotional response could characterized by a level of pleasure (valence) and intensity (arousal). As it reflects on the autonomic nervous system (ANS) activity, modern wearable wristbands can record non-invasively and during our everyday lives peripheral end-points of this response. While emotion recognition from physiological signals is usually achieved using supervised machine learning algorithms that require ground truth labels for training, collecting it is cumbersome and particularly unfeasible in naturalistic settings, and extracting meaningful insights from these signals requires domain knowledge and might be prone to bias. Here, we propose and test a deep-seeded clustering algorithm that automatically extracts and classifies features from those physiological signals with minimal supervision — combining an autoencoder (AE) for unsupervised feature representation and c-means clustering for fine-grained classification. We also show that the model obtains good performance results across three different datasets frequently used in affective computing studies (accuracies of 80.7\% on WESAD, 64.2\% on Stress-Predict and 61.0\% on CEAP360-VR).

\end{abstract}

\section{Introduction}
% def
Emotions play a significant role in the cognitive processes of the human brain, such as decision-making, learning and perception. Therefore, emotions affect our everyday life, and understanding it plays a vital role in understanding how we behave \cite{1antoine}.

Based on the circumplex model of affect, an emotion is best described as varying along two dimensions that relate to valence (unpleasant to pleasant) and arousal or activation (low to high intensity) \cite{russell1980circumplex}. The autonomic nervous system (ANS) is viewed as a critical component of the deployment of an emotional response, leading to peripheral physiological signals (such as heart rate, temperature or sweating responses) which cannot be intentionally controlled. Hence, measuring such signals may often provide more objective and reliable measures (compared to, for example, self-assessments) \cite{ans_review}. 

Modern wearable wristbands, such as the Empatica E4 \cite{empatica2015e4} (to note that it has been recently discontinued), can be used to record peripheral physiological signals (including blood volume pulse, or BVP); electrodermal activity, or EDA; and skin temperature, or TEMP) in a non-intrusive and convenient way – making it suitable for both lab-based and real-life settings.

In affective computing works, emotion recognition from physiological signals is usually achieved using supervised machine learning algorithms that require ground truth labels for training \cite{10_antoine} \cite{11_antoine} \cite{wesad} \cite{13_antoine}. However, collecting fine-grained labels (as opposed to single post-stimuli emotion labels) is often hard to implement and time-consuming. It is particularly not efficient (as it increases the risk of dropout and disengagement) when applying such emotion recognition algorithms to real-life settings with longitudinal data collection (e.g., continuous recordings over several days) \cite{zhang2022weakly}.

In this manuscript, we propose a novel approach by describing and assessing a novel deep-seeded clustering model for the characterisation of emotional responses derived from a single retrospective affective label. In brief, the suggested model combines an autoencoder (AE) – for unsupervised feature representation with c-means clustering – for fine-grained classification. The present work aims to provide the following relevant contributions to affective computing research:

\begin{itemize}
    \item Propose an end-to-end deep learning framework for emotion recognition – which omits feature extraction and requires no domain knowledge. For this, it takes (as input) raw physiological data collected from a wearable device; learns insightful feature representations (or embeddings); and infer, with minimal supervision (i.e., labels) emotions from those physiological measurements.
    \item Show reasonable to good model performance of our weakly-supervised framework on three validated datasets commonly used in the affective computing literature (WESAD \cite{wesad}; Stress-Predict \cite{spred}; and CEAP360-VR \cite{ceap_vr}). Our findings revealed, for subject-dependent classification, the following accuracies: 80.7\% accuracy on WESAD, 64.2\% on Stress-Predict and 61.0\% on CEAP360-VR. To note that these results were either similar or better than those achieved by the original studies – using supervised approaches. Furthermore, it is important to highlight that our framework has achieved such recognition accuracy and clustering quality without any hyper-parameter tuning – leading us to hypothesize that better performance results could be achieved through adequate optimization.
\end{itemize}

\section{Related work}
% def
In this section, we review related work in the field of machine learning for emotion recognition from physiological signals. We provide a brief overview of: (i) the concept of emotion; (ii) how it translates into physiological signals that we can easily and objectively measure; and (iii) how machine learning techniques have been used in the literature for its detection. To note that we will not be addressing neural emotional regulation or processing (i.e., brain processes controlling or modulating the emotional response – beyond the scope of this work); but rather the (self-) perceived or physically deployed (through physiological peripheral responses) emotional experiences.

\subsection{Emotion Theory, Elicitation and Assessment}
% def
Defining emotion in a universally accepted way is challenging. In general, affect is understood as a neurophysiological state that can be consciously experienced, but not necessarily focused on a specific object. Mood refers to a prolonged, low intensity feeling, while emotion is characterized by a brief, intense, and targeted experience \cite{fox2018perspectives}. In this work, we refer to emotions interchangeably with affect and consider the circumplex model of affect to define stress as a high-arousal and low-valence emotion \cite{russell1980circumplex}. 

A critical decision for a good experimental design (and for the validity or generalization of a scientific finding) is whether to conduct the study in a laboratory setting or in a real-life (or ecological) environment. In field studies, a key challenge is generating accurate labels; on the other hand, no design of affective stimuli is required, as different emotional states naturally occur in real-world settings. In contrast, when conducting lab-based studies, researchers can follow well-designed protocols which not only allow for replication, but also can facilitate obtaining high-quality labels, as the desired affective states can be induced using a carefully selected set of stimuli, and dedicated time slots can also be set aside for questionnaires; however, if these stimuli are not appropriate, the intended effects may not be achieved.

Most studies opt for lab-based settings, using images, videos, virtual reality (VR) environments, or stress-inducing protocols as stimuli \cite{slr}. Among those are \cite{wesad}, which combined videos and a social-evaluative and cognitive stress-inducing protocol, the Trier Social Stress Test (TSST), \cite{spred}, that applied stress-inducing tasks, including social-evaluative, cognitive and physical - the TSST, the Stroop Color Test and the Hyperventilation Provocation Test, and \cite{ceap_vr}, which used immersive VR for emotion elicitation. 

To ensure that the desired affective states are successfully evoked, self-assessment questionnaires are often used. These include the Self-Assessment Manikin (SAM), the State-Trait Anxiety Inventory (STAI), Positive and Negative Affect Schedule (PANAS) and self-developed questionnaires, for example, including rating the intensity of emotions or affect on a Likert scale \cite{slr} \cite{4_slr}. In real-life (field) studies, ecological momentary assessments (EMA) are often used, which consist of short sets of questionnaires that participants occasionally file to report their current affective state \cite{4_slr}. 

\subsection{Physiological Indicators of Affect} \label{section:physio}
% def

Physiological changes in the cardiovascular system, electrical conductance of the skin (electrical electrodermal activity, or EDA) and skin temperature have been associated with some affective states. Relevant for real-life experimental settings, some commercial wearable devices (to note that wearable devices can also be used in lab-based studies; but for such well-controlled studies, preference is given to more robust devices) often allow the recording of such physiological signals \cite{slr}\cite{4_slr}. As a consequence, several studies in the literature have been using these physiological signals for emotion recognition — including the Empatica E4  \cite{empatica2015e4} device used in our present work.

The Blood Volume Pulse (BVP), derived from photoplethysmography (PPG), monitors changes in blood volume within the capillaries and arteries and it is typically measured using non-invasive optical sensors that assess cardiovascular dynamics \cite{38} \cite{42}. While this sensor can be placed on various parts of the body, in the case of Empatica E4, it is embedded in its wristband. Heart rate information derived from BVP (computed through the inverse) is indicative of emotional states, with lower heart rates associated with relaxation and higher heart rates linked to heightened emotions such as stress, joy, or anger \cite{14}\cite{31}.

The Electrodermal Activity (EDA), also known as Galvanic Skin Response (GSR), is another (as the cardiovascular and temperature response) non-invasive peripheral measure of the autonomic nervous system (ANS) activity, being particularly related to the arousal state. For example, increased EDA has been associated to intense emotions like joy, anger, and stress \cite{12} \cite{31} \cite{42}. In brief, it reflects changes in the skin’s electrical properties (particularly its electrical conductance) due to sweat gland activity. To measure EDA, electrodes are placed on the skin, usually in the palms, fingers or in the wrist \cite{41},

Skin temperature, similar to EDA, is influenced by external factors linked to emotional responses; for instance, stress can cause muscle tension and blood vessel constriction, leading to a decrease in temperature \cite{42}. Despite being a useful emotional indicator, it is relatively slow to respond to changes, so a sampling rate of 1 Hz is generally sufficient \cite{4_slr}.

\subsection{Machine Learning for Emotion Recognition}

Various machine learning approaches have been used to infer emotional states, exploring both supervised and unsupervised techniques \cite{26}.

Supervised models learn to recognize emotions using a set of explanatory variables and respective labels (either self-reported emotional ratings or standardized stimuli/contexts well-known to elicit specific emotions) \cite{10_antoine} \cite{11_antoine} \cite{wesad} \cite{13_antoine}. Even though studies using these methods have achieved good performance in emotion recognition, the required ground truth labels are not only time consuming to obtain, but can also be imprecise as, on one hand, a specific stimuli may evoke a combination of emotions rather than a unique affective state, and, on the otber hand, self-reported psychological assessments are exposed to bias \cite{16_antoine} \cite{17_antoine} \cite{3_antoine} \cite{2_antoine}.

Regarding the input explanatory variables used for supervised learning, if there is domain knowledge, relevant physiological indicators as those introduced in section \ref{section:physio} can be identified and used for emotion recognition, and further techniques such as dimensionality reduction and feature selection can also be integrated to optimize the model's parameters. Alternatively, feature extraction (or model-free) methods often have shown to outperform models trained on manually selected statistical features. Some of these alternatives include autoencoders – which extract meaningful representations through the compression and reconstruction of unlabeled data; and deep neural networks – which can automatically extract complex patterns from multimodal signals, and thus be used to learn task-specific representations for each physiological signal in an end-to-end manner  \cite{20_antoine} \cite{21_antoine} \cite{38_antoine} \cite{39_antoine} \cite{40_antoine} \cite{41_antoine}. Nonetheless, some of these model-free methods can easily overfit training data, especially when using deep and complex structures \cite{45_weakly}.

Several studies have investigated the use of these model-free feature extraction methods for emotion recognition. In one study, stacked convolutional autoencoders were applied to unlabeled cardiac information (derived from electrocardiography) and EDA data to extract generalized latent representations for arousal classification, and outperformed fully supervised methods \cite{16_weakly}. Although this approach effectively handled the variability of different types of biosignals, it overlooked the complementary nature of multimodal data. In contrast, another study proposed CorrNet – a correlation-based emotion recognition algorithm that first extracted intramodal features (using separate convolutional autoencoders) and then computed covariance as well as cross-covariance between modalities to capture intermodal features \cite{15_weakly}. However, this unsupervised method did not incorporate supervised signals during pre-training, which may have negatively impacted its performance.

Other research employed sequential machine learning algorithms like Long Short-Term Memory (LSTM) networks to model the relationship between input signals and emotions \cite{4_weakly} \cite{10_weakly} \cite{11_weakly} \cite{12_weakly}. However, such algorithms require fine-grained emotion labels for training. When using post-stimuli labels to recognize these fine-grained emotions, the lack of information about which instances specifically correspond to these emotions can lead to overfitting and temporal ambiguity \cite{3_weakly} \cite{26_weakly} \cite{27_weakly}.

In recent years, unsupervised learning has gained attention as a promising solution to the scarcity of reliable labeled data in emotion recognition, with studies demonstrating its capacity to distinguish between valence and arousal. Some of these approaches used algorithms such as k-means, Gaussian Mixture Models (GMM) and Hidden Markov models (HMM) for binary classification \cite{22_antoine} \cite{23_antoine} \cite{24_antoine} \cite{25_antoine} \cite{26_antoine} \cite{27_antoine} \cite{28_antoine} \cite{29_antoine}. \cite{34_antoine} \cite{35_antoine} \cite{36_antoine}.

Although promising, unsupervised learning has several limitations. First, the lack of labels makes it difficult to interpret the predicted emotional states. For example, when clustering physiological data, it’s often unclear which cluster corresponds to which emotion, as the clusters do not map directly to emotional categories. This can be partially addressed using pseudo-labels or by seeding the algorithm with prior knowledge \cite{37_antoine}. Second, unsupervised models generally underperform compared to supervised approaches in emotion recognition tasks. Third, while popular clustering algorithms like k-means, DBSCAN, Affinity Propagation, and BIRCH assign data points to mutually exclusive clusters, emotional states are often ambiguous and overlapping. These methods do not allow for degrees of membership, which limits their ability to reflect the continuous nature of emotions \cite{russell1980circumplex}.

This manuscript introduces a novel deep-seeded clustering framework designed to address key limitations of both supervised and unsupervised approaches in emotion recognition. Our model employs a sequence-to-sequence autoencoder \cite{43_antoine} to extract features from physiological signals, which are jointly trained with a deep clustering algorithm. By seeding clusters with self-reported emotional assessments or contextual stimuli, the model effectively maps responses to the four quadrants of Russell’s circumplex model of affect. The proposed framework processes EDA, BVP, and skin temperature signals collected via the Empatica E4 wristband – and across multiple publicly available datasets – using a deep-seeded c-means clustering method for emotion recognition.

\section{Methodology}
% def
In this section, we introduce the theoretical framework upon which our proposed deep-seeded clustering model for emotion recognition is based. Our approach relies on three main components: pre-processing, representation learning and clustering. Theoretical and mathematical background regarding each of those steps is presented in the following subsections.

\subsection{Pre-processing} \label{sec:preprocessing}

A pre-processing step was applied prior to the learning stage, which included smoothing, min-max scaling, and resampling. A Savitzky-Golay filter with polynomial order 1 was used for smoothing, to mitigate errors and remove outliers. Min-max scaling to a 0-1 range was used to accelerate feature extraction and improve clustering results, leading to faster convergence and increased probability of finding a global minimum \cite{54_antoine}. As the Empatica E4 wearable device \cite{empatica2015e4} uses sampling frequencies of 4 Hz for the EDA and Skin Temperature signals and 64 Hz for the BVP signal, we opted for upsampling to the largest frequency, i.e., 64 Hz, to minimize information loss.

We applied sliding windows of 600 samples (roughly 10 seconds), overlapping with a step of 1 sample, and considered only those that were assigned to a single emotional context (self-reported or stimulus-based; see \ref{section:exp_Setup}). 

\subsection{AEs for representation learning}
Rather than using hand-crafted features, the proposed model uses an unsupervised algorithm to learn an efficient latent representation of the input data \cite{xu2016stacked}, particularly an autoencoder. In detail, an autoencoder consists of a multi-layer feed-forward neural network, which includes the input layer, a hidden layer and the output layer, without any directed loops or cycles \cite{ju2015deep}. It aims to minimize the discrepancy between input and reconstruction by learning an encoder and a decoder, yielding a set of weights $W$ and biases $b$ \cite{xu2014stacked}.

Defining the unlabeled training data as $\left \{ x^{(1)},...,x^{(n)} \right \}$, $n$ being the number of training samples, and the target values of this neural network as $\left \{ y^{(1)},...,y^{(n)} \right \}$, $x^{(i)} \in \mathbb{R}^p, y^{(i)} \in \mathbb{R}^p, \forall i$, an autoencoder thus aims at setting the target values to be equal to the inputs, using $y^{(i)}=x^{(i)},\forall i$ \cite{ju2015deep}. This is equivalent to saying that this algorithm tries to learn an approximation to the identity function, so as to output a reconstructed $\hat{x}$ that is similar to $x$, where this discrepancy to be minimized is described by an average sum-of-squares error term in the cost function, as given by:
\begin{equation} \label{eq:AE_Recerror}
L_{AE}=\frac{1}{n} \sum_{i=1}^{n} \bigg ( \frac{1}{2} \left \| y^{(i)}-h_{W,b}(x^{(i)}) \right \|^2 \bigg)
\end{equation}
\noindent where $L_{AE}$ corresponds to the reconstruction loss that the AE aims to minimize, and $h_{W,b}(x^{(i)})$ corresponds to the result of the output layer, for each input training pattern $x^{(i)}$, thus equivalent to its output reconstruction \cite{ju2015deep}. We can rewrite this equation as:
\begin{equation} \label{eq:AE_loss_antoine}
L_{AE}=\frac{1}{2} \bigg(  \frac{1}{n} \sum_{i=1}^{n} \left \| x^{(i)}-\hat{x}^{(i)} \right \|^2 \bigg)
\end{equation}

If there is structure in the data (if it is not completely random), by limiting the number of hidden units or enforcing a sparsity constraint, the network can then detect it; particularly, if the number of hidden units is much smaller than the input and output layer, the network is forced to learn a compressed representation of the input, thus allowing to identify existing correlations in the input features, as mentioned \cite{ju2015deep}. 

While the vanilla autoencoder consists of this simpler multi-layer feed-forward network, more sophisticated encoder and decoder layers can be designed and implemented in its place, particularly long short-term memory networks (LSTMs) and gated recurrent units (GRUs), which are frequently employed in time-series classification problems \cite{43_antoine}. Unlike standard feed-forward neural networks, these have feedback connections, allowing them to exploit temporal dependencies across sequences of data \cite{12_weakly}\cite{hochreiter1997long}. 

LSTMs were originally designed to handle the issue of vanishing or exploding gradients that can occur when training traditional recurrent neural networks, and they are particularly effective for learning and predicting on sequence data due to their ability to retain memory over long sequences \cite{hochreiter1997long}. GRUs, on the other hand, use a gating mechanism which is very similar to LSTMs, but are faster to train as they require fewer parameters and only two gates: the update gate ($u_t$), which tunes the update speed of the hidden state, and the reset gate ($r_t$), which decides how much of the past information to forget by resetting parts of the memory \cite{cahuantzi2023comparison}\cite{cho2014properties}. 

Since our goal is to  to perform feature extraction and signal reconstruction of sequential data, we use sequence-to-sequence (S2S) autoencoders integrating encoder and decoder GRU layers, to efficiently reconstruct
the EDA, BVP and Skin Temperature signals while minimizing the number
of parameters \cite{43_antoine}. Equation \ref{eq:AE_loss_antoine} thus becomes:

\begin{equation} \label{eq:loss_AE_ref}
L_{AE}=\frac{1}{2} \bigg( \frac{1}{n} \sum_{t=0}^{T-\delta-1} \left \| x_{t}-\hat{x_t} \right \|^2 \bigg)
\end{equation}
\\ where $\delta$ is the window size (in our case, $\delta=600$ - see \ref{sec:preprocessing}), and $x_t=x_{t:t+\delta}, \forall t>=0$.

\subsection{Clustering}
In line with our motivation to develop a model which can be used in real-life settings, requiring minimal labeling of the data, the model we propose combines the AE architecture with c-means - an extension of the popular unsupervised learning algorithm k-means \cite{46_antoine}.

The goal of k-means is to cluster the different data points in meaningful groups such that each data point is closer to the nearest point (to which the distance should ideally be small) in the same cluster than to all the ones in the remaining clusters, resembling a classification problem (where each cluster can be seen as a different category or, in our case, a different affective state/emotion), but without the need for any expert labeling. Formally, this corresponds to a global optimization problem where the cost function to be minimized is given by \cite{jain2010data}:
\begin{equation}  \label{eq:loss_kmeans}
    L_{km}=\sum_{i=1}^{K} \sum_{x_t\in T_i} \left \| x_t-\mu_i \right \|^2
\end{equation}
where $T= \left \{x_1,...,x_n  \right \} $ corresponds to the feature vectors to be clustered (the training set), $T_i$ to the ones assigned to the $i$-th cluster, $\mu_i$ to the centroid of each $i$-th cluster and $K$ to the total number of clusters. Assignment to a cluster and centroid update are thus iteratively performed in K-means, in order to find K centroids (corresponding to the center of the clusters) such that after assigning each vector to the nearest center, the sum of squared distances from the centers are minimized \cite{jain2010data},\cite{leung2001representing}, which we be formalized as:
\begin{equation} \label{eq:dataassignment}
    x_t \in T_j: j = arg \ \underset{i }{min} \left \| x_t-\mu_i \right \|^2
\end{equation}
\begin{equation} \label{eq:centroidupd}
\mu_j= \frac{1}{\#T_j}\sum_{x_t \in T_j} x_t
\end{equation}
\\
We can rewrite equation \ref{eq:loss_kmeans} as:
\begin{equation} 
    L_{km}=\sum_{t=0}^T \sum_{k=1}^{K} s_{tk} \left \| x_t-\mu_k \right \|^2
\end{equation}
\\
defining $s_{tk}$ as the assignment vector $s_t \in \{0,1\}$ such that:
\begin{equation} 
s_{tk}=\begin{cases}
      1, \; \text{if} \; k =  arg \ \underset{i }{min} \left \| x_t-\mu_i 
      \right \|^2 \\
      0, \; \text{otherwise}
    \end{cases}\,.
\end{equation}

C-means generalises k-means by assigning data points to
fuzzy clusters instead of mutually exclusive clusters. According to \cite{46_antoine}, observation $x: x\in T_i$ belongs to cluster $i$ with degree:
\begin{equation} \label{eq:u_tk}
u_{tk}= \frac{exp(-\frac{2}{\gamma} \frac{\left \| x_t-\mu_k \right \|^2}{1+\left \| x_t-\mu_k \right \|^2})}{\sum_{l=1}^K exp(-\frac{2}{\gamma} \frac{\left \| x_t-\mu_l \right \|^2}{1+\left \| x_t-\mu_l \right \|^2})}
\end{equation}
\\
with $\gamma > 0$. The higher is $u_{k}$, the higher is the probability that $x$ belongs to cluster $k$. Moreover, centroid $c_k$ of fuzzy cluster $k$ is defined by:
\begin{equation} \label{eq:mu_k}
\mu_{k}= \frac{\sum_{t=0}^T d_{tk}u_{tk}x_t }{\sum_{s=0}^T d_{sk}u_{sk}}
\end{equation}
\\
where
\begin{equation} 
d_{tk}= \frac{\left \| x_t-\mu_k \right \|+2}{(\left \| x_t-\mu_k \right \|+1)^2}
\end{equation}
\\
C-means’ estimation steps are the same as those of k-means. Thus, the algorithm works by successively updating the assignment vectors from equation \ref{eq:u_tk} and the centroids from equation \ref{eq:mu_k}, minimizing the loss:
\begin{equation} 
    L_{cm}=\sum_{t=0}^T \sum_{k=1}^{K} u_{tk} \left \| x_t-\mu_k \right \|^2
\end{equation}

\subsection{Deep-seeded clustering} \label{deepseededclustering}

Considering that there are multiple solutions to the optimization problem of minimizing the AE reconstruction loss in equation \ref{eq:loss_AE_ref}, additional criteria can be introduced to ensure that the learned latent representation is the most suitable to the problem at hand. In this case, since our ultimate goal is to perform emotion recognition through clustering, the features learned at this step should also maximize the quality of the clusters found later on through c-means. In our proposed method, we thus jointly estimate the feature representation and clustering models by setting our loss function to be the sum of both parts:
\begin{equation} \label{eq:bigL}
L= L_{AE} + L_{cm}
\end{equation}

\begin{figure}[h]
    \centering
    \includegraphics[scale=0.25]{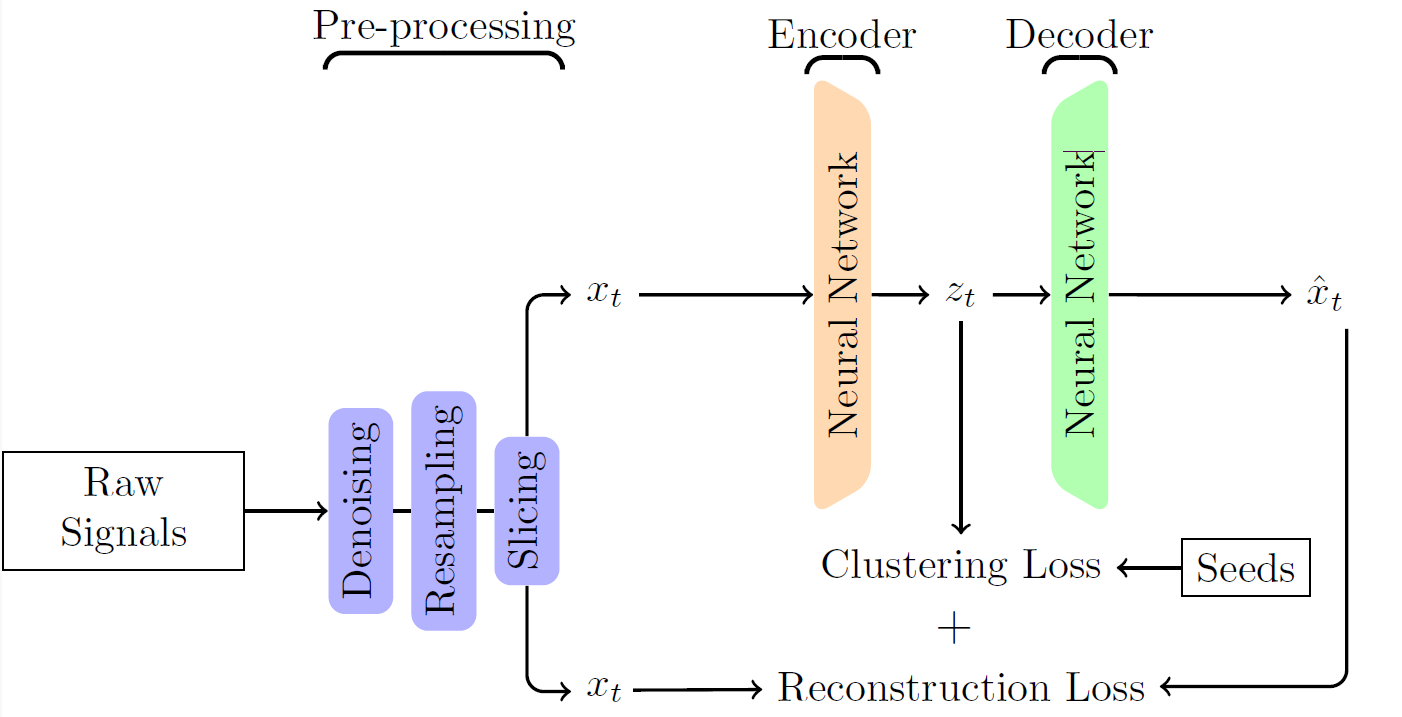}
    \caption{Diagram illustrating the proposed model architecture.}
    \label{fig:diagrama_modelo}
\end{figure}

The output of the hidden layer of the autoencoder is thus given as input for c-means clustering, so as to label the resulting $D$-dimensional encoded vectors (in our case, considering $D=30$, see \ref{section:exp_Setup}), while the parameters of the two networks are updated together, at each training epoch of the full model, through backpropagation. The proposed model architecture is depicted in Figure \ref{fig:diagrama_modelo}.

Centroid initialization can be done by simply picking a set of random feature vectors, but the algorithm can converge to different solutions depending on the initialization. Rather than following this approach, we initialized all data points at training time according to contextual or else self-reported data (see \ref{section:exp_Setup}). Indeed, considering that c-means is an unsupervised model, the resulting clusters are not meaningful unless a seeding method is used, as they only provide information about which points belong to the same group, but not about which group corresponds to each class/emotion \cite{37_antoine}\cite{50_antoine}.

In the self-reported case, if individuals assess that an emotion $k^*$ is dominant from $T_1$ to $T_2$, physiological signal windows $(x_t)_{t=T_1}^{T_2}$ are likely to reflect it. Thus, for c-means clustering, we initialize the assignment probabilities as $u_{tk*}=1, u_{tk}=0, \forall t \in [T1,T2], k \neq k^*$. Thereby, emotion $k^*$ defines the pseudolabel of cluster $k^*$. Since this is a retrospective label, one should expect that not all time points within that interval would reflect that emotion $k^*$, or at least not with the same intensity. At each iteration of the model, the assignment probabilities are thus adjusted so as to minimize the combined loss from AE and c-means, so that the final probabilities are heterogeneous and reflect cluster membership, i.e. the degree to which each emotion is present. The same reasoning applies to the contextual approach, where self-reported labels are replaced by the stimuli.

This procedure of centroid initialization works as seeding the model, as, by initializing the different clusters in this way, we can claim that the data points (in this case, the different time windows) assigned to the cluster of each respective centroid (which is also updated throughout time) also belong to that same emotion, and thus extract metrics of semi-supervised model performance (predicted against ground truth).

\section{Datasets} \label{datasets}
% def

To evaluate the accuracy and robustness of the presented method, three distinct datasets from the literature were considered (WESAD \cite{wesad}, Stress-Predict \cite{spred} and CEAP-360VR \cite{ceap_vr}).

In the case of the WESAD dataset, the goal was to elicit in 17 healthy participants three distinct affective states — neutral, stress (induced with the The Trier Social Stress Test, or TSST — a widely used method in psychology and neuroscience to make participants feel stressed in a controlled and safe way) and amusement (using a set of funny video clips). Although the dataset also includes two guided meditation periods, these were intended to facilitate de-excitation following the stress and amusement conditions. As the emotional response during meditation likely reflects a blend of the primary states rather than a distinct affective category, these periods were not treated as separate affective states for our analysis.

The full WESAD protocol lasted approximately two hours and included two variations: in one, the stress condition was placed between the two meditation periods; in the other, the amusement condition took that position. These variations were alternated among the participants to counterbalance potential order effects. At the end of each condition, participants reported their emotional state using the Self-Assessment Manikin (SAM) questionnaire  \cite{SAM}. However, as some conditions may not have elicited the intended emotional responses  — e.g., a participant might rate the stress condition as high valence and high arousal (or amusement-like) instead of low valence and high arousal (or stress-like) — different labelling strategies can be considered. One approach is to use the contextual label based on the experimental condition (e.g., “stress”), while another is to use the self-reported emotional label reflecting the participant’s actual experience (e.g., “amusement”). However, relying solely on self-reports often leads to data collapsing into the same emotional quadrant of the affect circumplex model, resulting in pseudo-labeled datasets with limited class diversity (e.g., only one or two emotional classes). Thus, contextual data was used, and the results of the classification task using the three stimuli can be found in section \ref{section:results}.

\begin{comment}
\begin{figure}[h!]
    \centering
    \includegraphics[scale=0.8]{pictures/wesad_protocol.jpg}
    \caption{The two different versions of the WESAD's study protocol \cite{wesad}, interchanged between different subjects to avoid effects of order. The red/dark boxes refer to filling in self-reports. In total, the study had a duration of about two hours.} 
    \label{fig:wesad_protocol}
\end{figure}
\end{comment}

In the Stress-Predict dataset, 35 healthy volunteers participated in a series of tasks designed to induce stress (the Stroop Color Test, the Trier Social Stress Test, and the Hyperventilation Provocation Test), each separated by a rest period. Participants also completed questionnaires aimed at inducing stress levels comparable to those encountered in daily life. The entire protocol lasted approximately one hour and was designed to elicit two distinct affective states: neutral (baseline) and stress. The results of the classification task using these two contextual “stimuli” labels are presented in Section \ref{section:results}.

\begin{comment}
Unlike WESAD, the SAM questionnaire was not employed; thus, only the contextual "stimuli" classification was considered for this dataset.

\begin{figure}[h!]
   \vspace{-10pt}
    \centering
    \includegraphics[scale=0.35]{pictures/spred_protocol.jpg}
    \caption{The Stress-Predict study protocol \cite{spred}, where the tasks meant to induce stress (Questionnaires, Stroop, Trier Social Stress Test and Hyperventilation Test) are interleaved with rest periods. In total, the study had a duration of about one hour.} 
    \label{fig:spred_protocol}
\end{figure}
\end{comment}

In the CEAP360-VR dataset, 32 healthy volunteers completed a VR task lasting approximately one hour. Using a commercial VR headset, participants watched eight validated 360\textdegree affective video clips selected from \cite{li14} — two representing each quadrant of the circumplex model of affect. While viewing the clips, participants annotated their perceived valence and arousal using a joystick controller (moving it along a spatial representation of the four-quadrant model). These annotations were recorded during and after each video, although only retrospective ratings were considered in this study. 

As each video of the CEAP360-VR dataset was clipped to a 60-second duration, the number of usable data points was limited, making multi-class classification infeasible. To address this, the four emotional quadrants were regrouped into two classes, transforming the task into a binary classification problem: stress vs. not stress, based either on the contextual “stimuli” labels or the retrospective self-reported “emotional” labels. The results for both approaches are presented and discussed in Section \ref{section:results}.

\begin{comment}
\begin{figure}[h!]
    \centering
    \includegraphics[scale=0.45]{pictures/ceapvr_protocol.jpg}
    \caption{The CEAP360-VR study protocol \cite{ceap_vr}. In total, the study had a duration of about one hour.} 
    \label{fig:ceapvr_protocol}
\end{figure}
\end{comment}

%WESAD: B|S|R|M|A|M (1) / B|A|M|S|R|M (2)
%Urban-Spaces: UC/UNC/NC/NNC random order

\begin{figure}[h!]
   \vspace{10pt}
%\smallskip
   %\centering
    \subfigure[]{
     \includegraphics[width=0.48\textwidth]{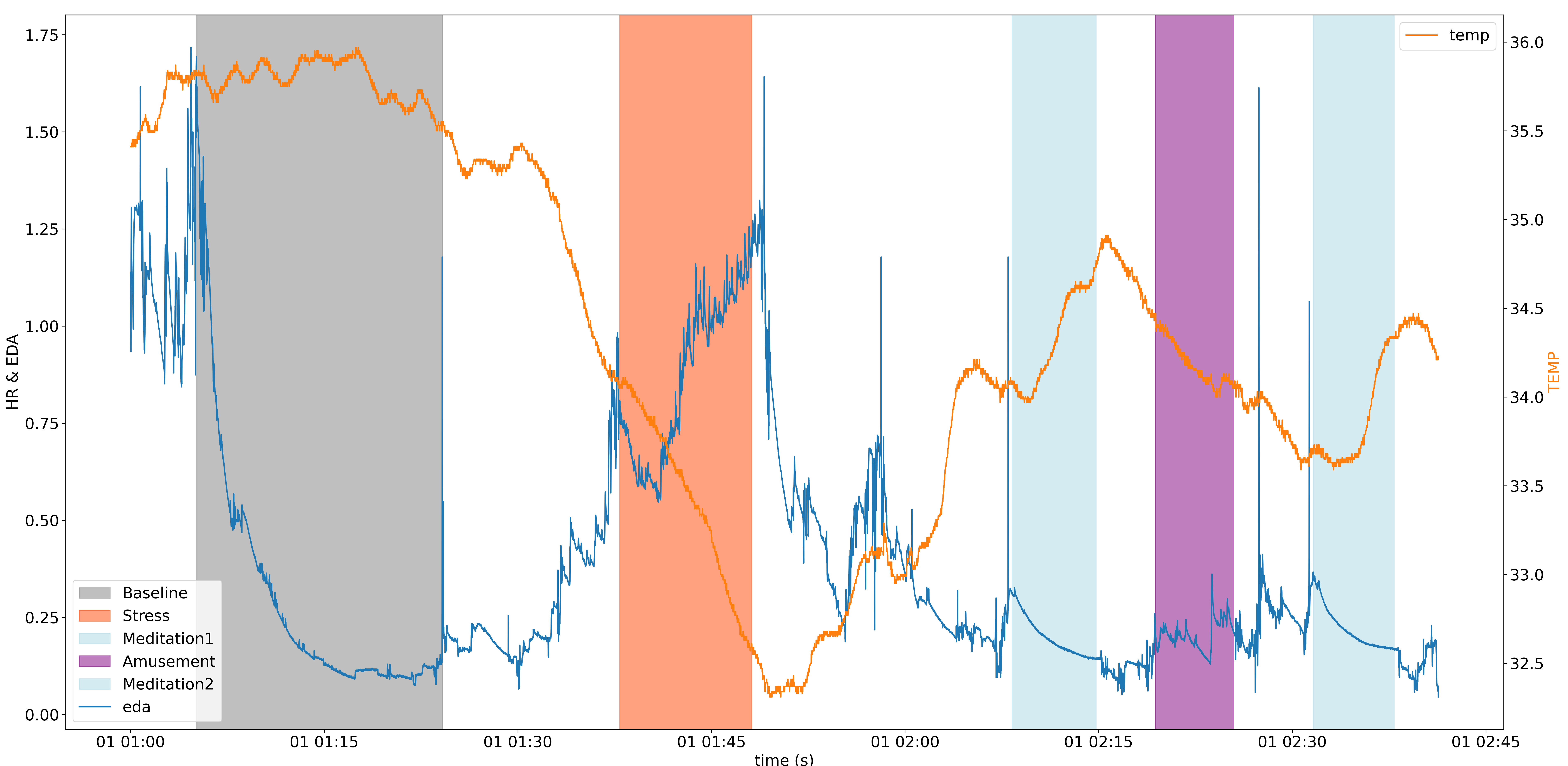}
     \vspace{10pt}
          \label{fig:wesad_edaretrosp}
} \\[10pt]
    \subfigure[]{
    \includegraphics[width=0.48\textwidth]{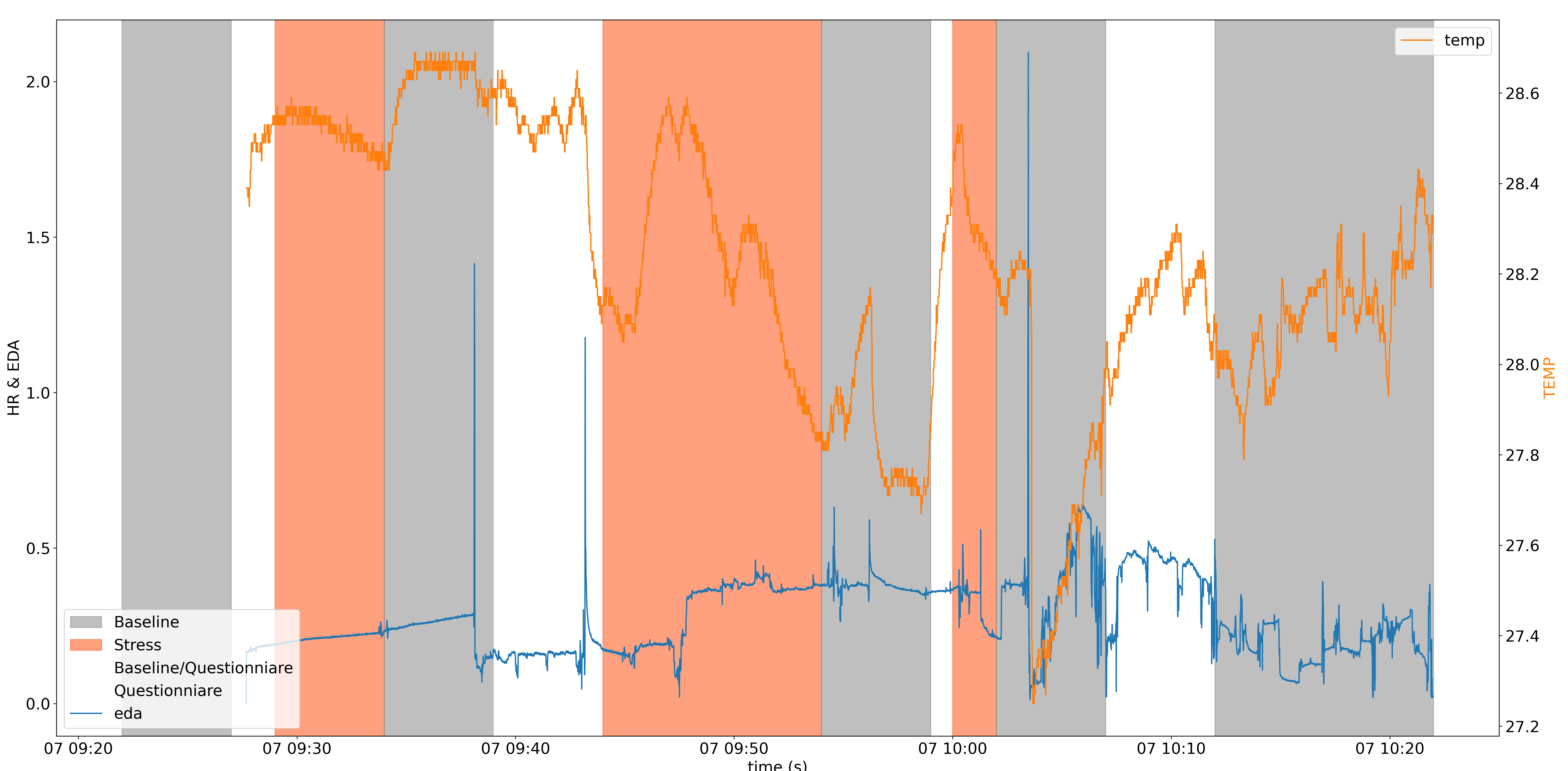}
         \vspace{10pt}
    \label{fig:spred_edaretrosp}
    } \\[10pt]
    \subfigure[]{
     \includegraphics[width=0.48\textwidth]{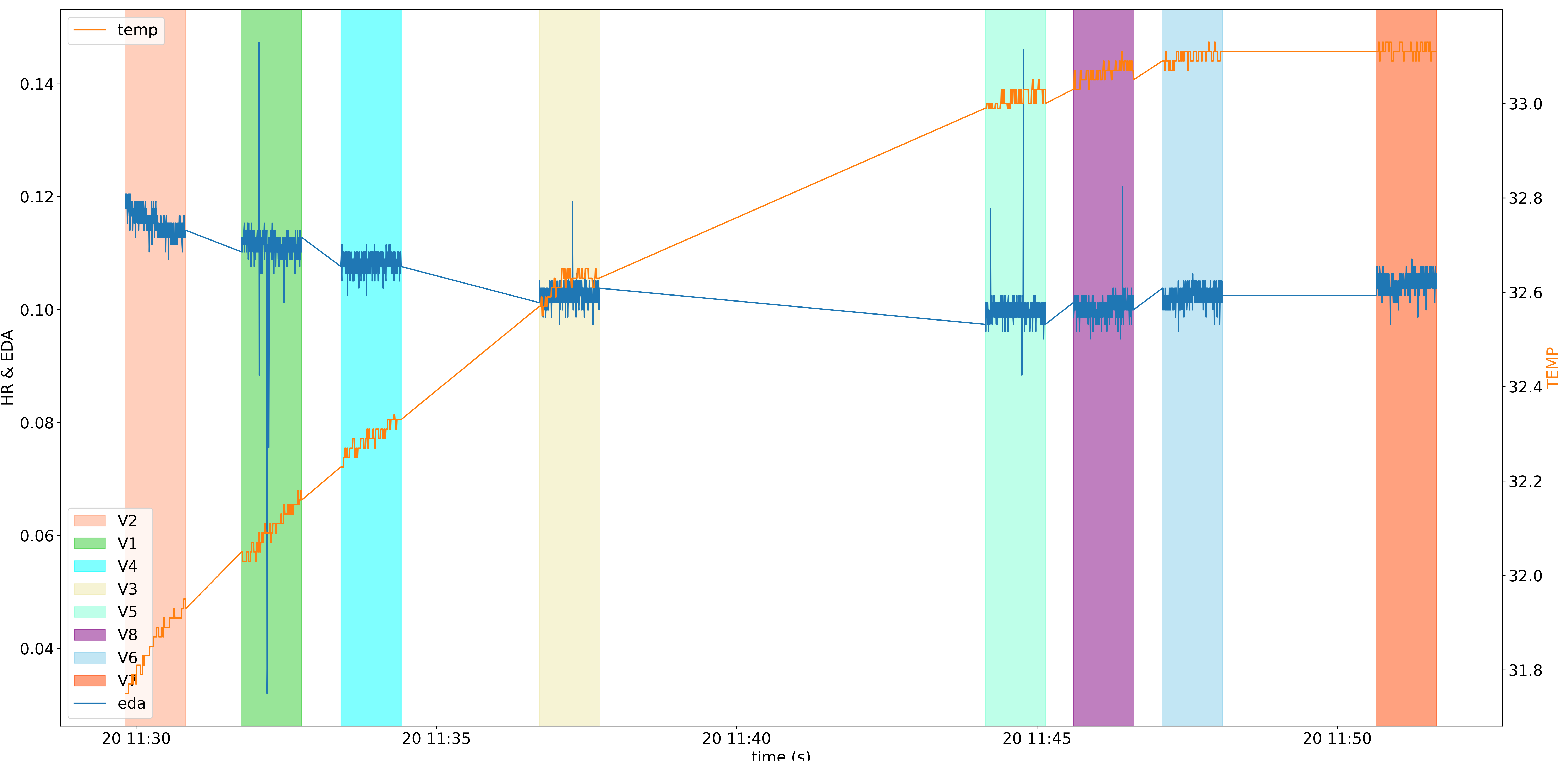}
          \vspace{10pt}
          \label{fig:ceapvr_edaretrosp}} \\[10pt]
   \caption[Sample images for each dataset.] {\label{fig:dataset_imgs} Sample images for each dataset: \subref{fig:wesad_edaretrosp} WESAD, \subref{fig:spred_edaretrosp} Stress-Predict, and \subref{fig:ceapvr_edaretrosp} CEAP360-VR.}
   \end{figure}

A sample image of the timeline and collected physiological signals for each dataset is presented in Figure \ref{fig:dataset_imgs} while summary information can be found in Table \ref{table:subjects}. 

%\begin{comment}
%\end{comment}

%\begin{comment}
\begin{table}[h!]
\begin{center}
\caption[Characteristics of each dataset]{Characteristics of each dataset \\ Format: Mean (Standard Deviation)}
\begin{tabular}{ |c|c|c|c|}
\hline
Dataset        & WESAD         & Stress-Predict        & CEAP360-VR         \\
\hline
\# Subjects     & 15         & 35         & 32         \\
Age               & 27.5 (2.4)  & 32 (8.2) & 25 (4.0) \\
\% of Males & 80.0 \%        &  28.6 \%      & 50.0 \%      \\
Classification Task              &   Multi-class: 3 &  Binary: 2  &  Binary: 2 \\
\hline
\end{tabular}
\label{table:subjects}
\end{center}
\end{table}
%\end{comment}

\section{Model Application} \label{section:results_discussion}
In this section, we present and discuss the model's implementation details and classification performance results on each dataset, considering a within-subject approach.

\subsection{Experimental setup} \label{section:exp_Setup}

The deep-seeded clustering model used in this work was trained with the library \textit{Pytorch} for \textit{Python} \cite{pytorch} \cite{python}. 

The encoder and decoder used in the autoencoder model were formed by two gated recurrent units (GRU) layers each, followed by normalisation layers to reduce the training time and improve clustering \cite{54_antoine}\cite{55_antoine} of the latent vectors and a fully connected layer with linear activation at the end of the decoder layer to restore the input dimensions. A hidden size of 30 was used for the autoencoder, i.e., a 30-dimensional embedding vector extracted from the encoder layer was given as input to c-means, and the joint model was trained over 100 epochs.

A within-subject approach was considered for the model. For each subject, the dataset was split in 10 folds for cross-validation, so that each rotating split was used for testing while the remaining 90\% of data was used for training. Two approaches were considered for this split: either using sequential, or non-sequential folds. In the former, the training dataset was downsampled by a factor of 10 - both to avoid overfitting and reduce training time -, while in the latter a downsampling factor of 2000 was considered - to minimize any partial overlap between the training and test sets.

The model was initialized by pre-training during a single epoch where only the autoencoder's reconstruction loss was considered, before proceeding to training the deep-seeded clustering model as a whole. The model's parameters were fixed at $\gamma=0.1$ and learning rates of $\eta=5*10^{-5}$ and $\eta=1*10^{-6}$ for training and pre-training, respectively. Model performance was estimated using the accuracy, precision and recall metrics:

%\begin{equation} \label{eq:f1score}
%F1-score=\frac{TP}{TP+(1/2)*(FP+FN)}
%\end{equation}
\begin{equation} \label{eq:accuracy}
Accuracy=\frac{TP+TN}{TP+TN+FP+FN}
\end{equation}
\begin{equation} \label{eq:precision}
Precision=\frac{TP}{TP+FP}
\end{equation}
\begin{equation} \label{eq:recall}
Recall=\frac{TP}{TP+FN}
\end{equation}

\noindent where $TP$, $TN$, $FP$ and $FN$ stand for the number of true positives, true negatives, false positives and false negatives, respectively. The experimental details for implementation, i.e., the model parameters, are summarized in Table \ref{table:model_params}.

\begin{table}[h!]
\begin{center}
\caption[Model Parameters]{Model Parameters}
\begin{tabular}{ |c|c|c|c|}
\hline
Parameter & Definition & Value \\
\hline
$D$ & Embedding size & $30$ \\
$\delta$ & Sequence length &  $600$ samples \\
$Epochs$ & Training epochs &  $100$ \\
$\gamma$ & Tradeoff parameter &  $0.1$    \\
$\eta_{ training}$ & Learning rate (training) & $5*10^{-5}$   \\
$\eta_{ pre-training}$ & Learning rate (pre-training) &  $1*10^{-6}$    \\

\hline
\end{tabular}
\label{table:model_params}
\end{center}
\end{table}

As previously mentioned, seeding (for both centroid initialization and ground truth labeling) was based on either context or self-reported data, depending on the dataset. The results of those different experiments are presented in section \ref{section:results}.

\subsection{Results and Discussion} \label{section:results}
In this section, we present model performance results, considering the accuracy, precision and recall metrics, as well as silhouette scores. To show that the model did not overfit to the training data, both training and test set results are shown. Both sequential and non-sequential splits were considered for training/testing the model. Since the second approach led to better performance across all datasets, we focus on it, although accuracy results for the sequential split are also presented. Results using the non-sequential approach are summarized in Table \ref{table:model_res}, for the three different datasets considered.

\begin{table}[h!]
\begin{center}
\caption[Classification Results]{Average classification results for each dataset (Macro)}
\begin{tabular}{ |c|c|c|c|}
\hline
Dataset        & WESAD         & Stress-Predict        & CEAP360-VR         \\
\hline
Accuracy     & 80.7 \%         & 64.2 \%           & 61.0 \%         \\
Precision              & 82.0 \%   & 65.2 \% & 55.2 \% \\
%Precision (Micro) & 80.7 \%        & 64.2 \%      & 50.0 \%      \\
Recall              & 84.7 \%  & 64.1 \% &  57.8 \%  \\
%Recall (Micro) & 80.7 \%        &  64.2 \%       & 50.0 \%      \\
%F1-Score               & 79.5 \%   & 62.9 \% & 50.9 \%   \\
%F1-Score (Micro) & 80.7 \%         &  64.2 \%     & 50.0 \%      \\
\hline
\end{tabular}
\label{table:model_res}
\end{center}
\end{table}

\hfill

\subsubsection{\textbf{WESAD}} \label{section:res_wesad}
\begin{figure}[h]
    \centering
    \subfigure{
        \includegraphics[width=0.38\textwidth]{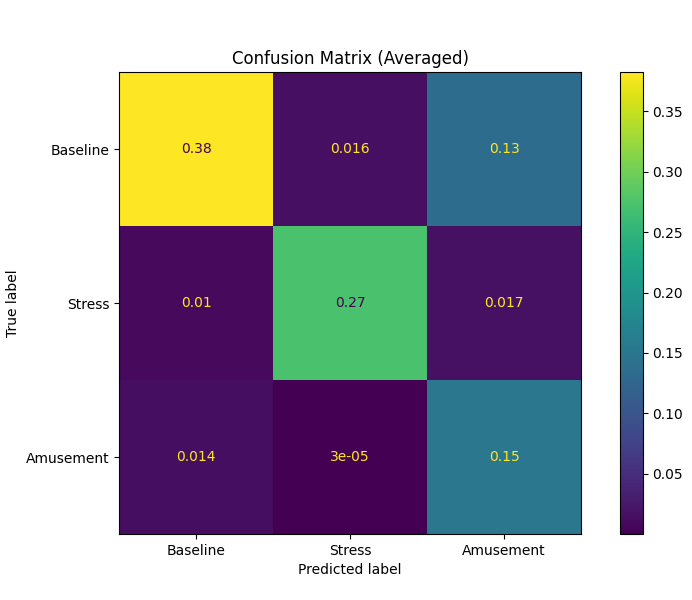}}
    \caption{Confusion matrix for WESAD using non-sequential 10-fold CV (within-subject), averaged across all subjects, considering contextual "stimuli" labels for seeding.}
    \label{fig:conf_wesad_NOTseq}
\end{figure}

Contextual data was used for seeding in WESAD, and the three different classes were considered: baseline, stress and amusement. The 3-class classification problem attained an accuracy of 79.3\% using sequential CV and 80.7\% using non-sequential CV, showing good performance -  similar to the 80.3\% obtained in the original paper \cite{wesad}. The confusion matrix for the non-sequential approach (averaged across subjects) is shown in Figure \ref{fig:conf_wesad_NOTseq}, illustrating how, in WESAD, stress and amusement are easily distinguished from the remaining contexts, while the baseline context is more frequently misclassified as amusement.

\begin{comment}
\begin{figure}[h!]
   \vspace{0pt}
%\smallskip
   %\centering
    \subfigure[]{
     \label{fig:conf_wesad_seq}
     \includegraphics[width=0.23\textwidth]{pictures/conf_wesad.png}} % \\[-1pt]
    \subfigure[]{
     \label{fig:conf_wesad_NOTseq}
     \includegraphics[width=0.23\textwidth]{pictures/conf_wesad_NOTseq.png}} % \\[-5pt]
\vspace{0pt}
   \caption[Confusion matrices for WESAD using 10-fold CV (within-subject).] {\label{fig:conf_wesad} Confusion matrices for WESAD using sequential \subref{fig:conf_wesad_seq} and non-sequential  \subref{fig:conf_wesad_NOTseq} 10-fold CV (within-subject), using contextual "stimuli" labels for seeding.}
   \end{figure}
\end{comment}

\begin{figure}[h]
    \centering
    \includegraphics[width=0.48\textwidth]{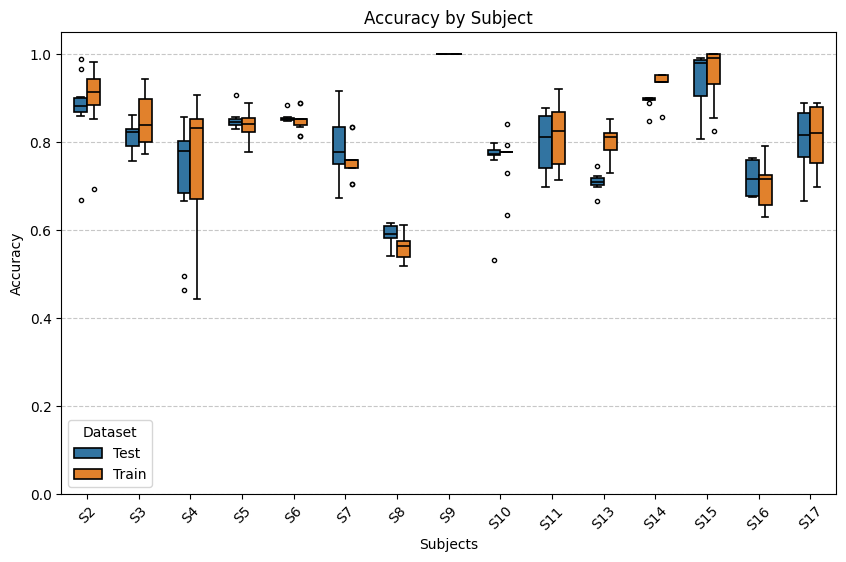}
    \caption{The non-sequential 10-fold CV accuracy for individual subjects of the WESAD dataset in the test (blue) and train (orange) sets.}
        \label{fig:acc_wesad_NOTseq}
\end{figure}

Figure \ref{fig:acc_wesad_NOTseq} illustrates the accuracy results in WESAD for individual subjects. It shows that, apart from subject S8, both the median accuracy and the 25\% quartile (across folds) is always above 60\%, and most often above 75\%, at testing. The model performance results are similar at training and testing as the two ranges intersect and there is no significant difference between median results for training versus testing. Thus, the model shows no signs of overfitting.

\begin{figure}[h]
    \centering
    \includegraphics[width=0.48\textwidth]{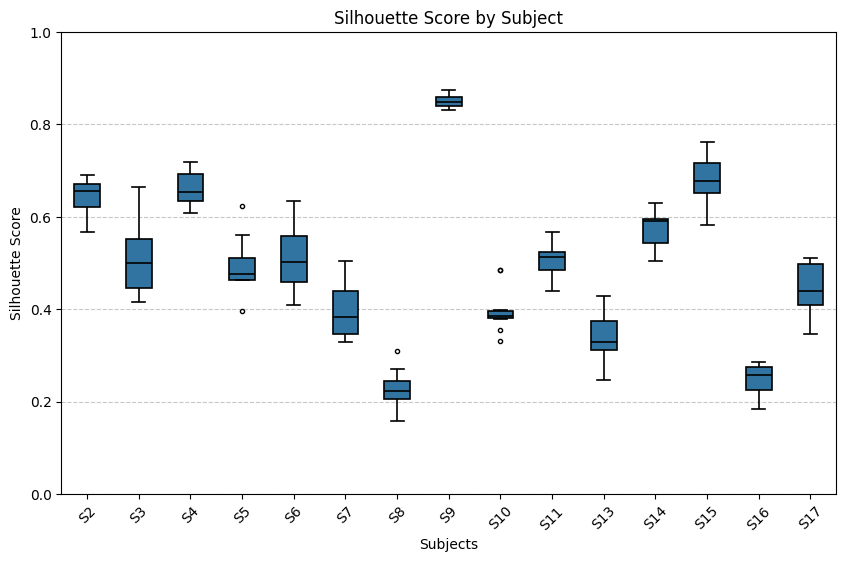}
    \caption{The non-sequential 10-fold CV silhouette scores for individual subjects of the WESAD dataset.}
        \label{fig:silhouettes_Wesad}
\end{figure}

This result is also in line with the silhouette score results presented in Figure \ref{fig:silhouettes_Wesad}, which illustrate how the model was able to extract meaningful and distinct clusters (one for each context) for most subjects (mean and median silhouette scores above 0.5), except S8 and S16, for which the quality of the clusters as given by this metric is extremely low.

\begin{figure}[h]
    \centering
        \subfigure{
    \includegraphics[width=0.38\textwidth]{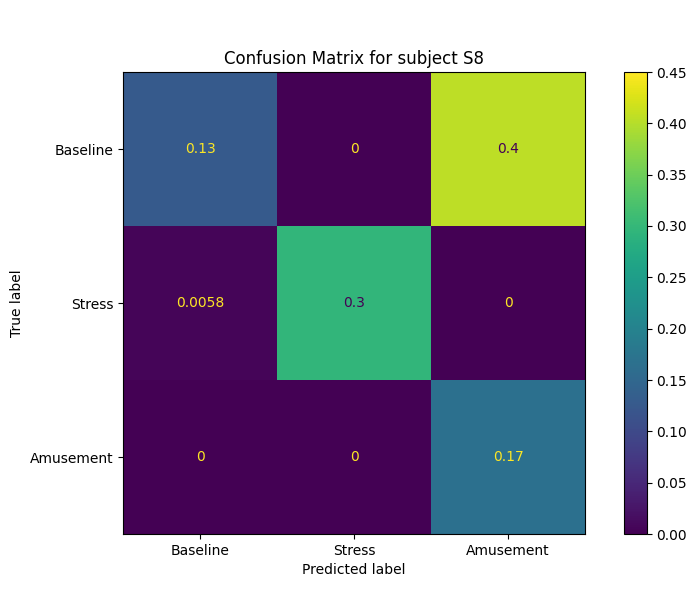}}
    \caption{Confusion matrix for subject S8 in WESAD, considering contextual "stimuli" labels for seeding.}
    \label{fig:conf_wesad_bad}
\end{figure}

To inspect the reason behind the model behaving differently for subject S8, its confusion matrix is presented in Figure \ref{fig:conf_wesad_bad}. This illustrates how, for subject S8, the baseline context is even more frequently misclassified as amusement than baseline, explaining the poor silhouette scores obtained for this participant. This is in line with the participant's self-reported data, where amusement and baseline were not only assigned to the same quadrant of the circumplex model of affect ("calmness"), but were also, in fact, scored with the exact same valence and arousal levels (valence: 7, arousal: 3) - unlike stress, which was scored with a lower valence and higher arousal level (valence: 5, arousal: 7). 

\begin{figure}[h]
    \centering
        \subfigure{
    \includegraphics[width=0.38\textwidth]{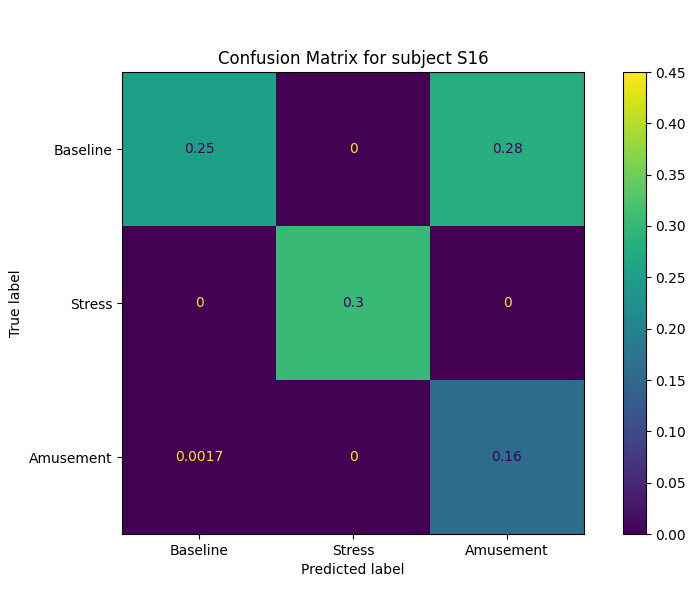}}
    \caption{Confusion matrix for subject S16 in WESAD, considering contextual "stimuli" labels for seeding.}
    \label{fig:conf_wesad_bad_s16}
\end{figure}

The low silhouette scores obtained for S16 are also in line with the self-reported data. Indeed, only two affective states were reported by the participant as well, with amusement and baseline falling once again into the same quadrants. Since half of the data in the baseline was still correctly identified, model accuracy remained high (above 60\%), as shown in Figure \ref{fig:conf_wesad_bad_s16}. 

\begin{comment}
\begin{figure}[h!]
    \centering
    \includegraphics[scale=0.4]{pictures/res_wesad.png}
    \label{fig:res_wesad}
    \caption{The non-sequential 10-fold CV classification results for the WESAD dataset: accuracy, precision (macro/micro), recall (macro/micro), and F1-score (macro/micro).}
\end{figure}

    \begin{figure}[h!]
    \centering
    \includegraphics[scale=0.5]{pictures/conf_wesad.jpg}
    \caption{Confusion matrix for WESAD using 10-fold CV (within-subject).}
\end{figure}
    \textcolor{gray}{- seeding:
1) only context agg (Stress vs. NOT Stress) or 4 (Neutral vs. Stress vs. Amusement vs. Meditation)
2) self-reported (quadrants from SAM: Neutral vs. Stress vs. Amusement)
3) pop (POPULATION self-reported quadrants from SAM: Neutral vs. Stress vs. Amusement) 
- discuss results}
\end{comment}

\hfill

\subsubsection{\textbf{Stress-Predict Dataset}} \label{sec:res_spred}
Contextual data was also used for seeding in Stress-Predict, and two different classes were considered: stress and baseline. The accuracy obtained for this binary classification problem was close to 60.5\%  using sequential CV and 64.2\% using non-sequential CV, showing reasonable performance - close to the 67\% obtained in the original paper \cite{spred}.

\begin{figure}[h]
    \centering
    \includegraphics[width=0.38\textwidth]{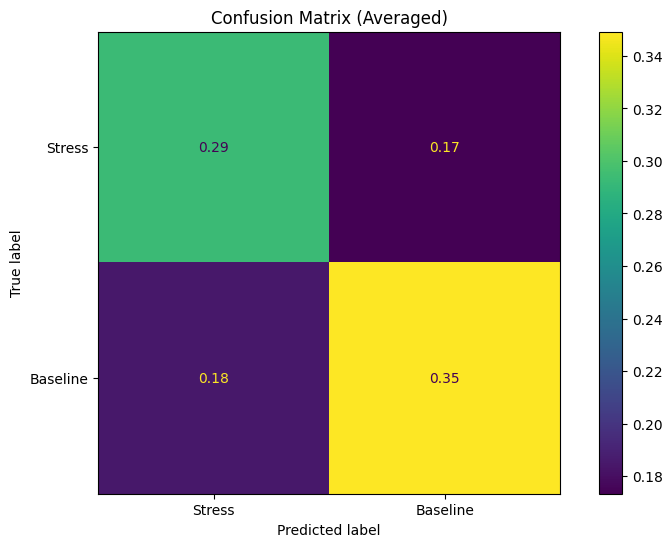}
    \caption{Confusion matrix for Stress-Predict using non-sequential 10-fold CV (within-subject), considering contextual "stimuli" labels for seeding,}
        \label{fig:conf_spred_NOTseq}
\end{figure}

The confusion matrix (averaged across subjects) is shown in Figure \ref{fig:conf_spred_NOTseq}, illustrating that, while baseline contexts are more easily detected, a larger fraction of stress contexts are misclassified as baseline (not stress). Notably, three different stress induction procedures were used in this dataset, including TSST (similar to WESAD) but also the Hyperventilation Provocation Test and the Stroop Test, which might not have been as effective. 

The fact that different durations were considered for the three stress induction paradigms explains why the non-sequential leads to better performance results compared to the sequential approach. Indeed, in the non-sequential approach, data points from all stress contexts are used for training, while in the sequential approach this distribution is unbalanced, and can lead to out-of-distribution testing.

\begin{comment}
\begin{figure}[h!]
   \vspace{0pt}
%\smallskip
   %\centering
    \subfigure[]{
     \label{fig:conf_spred_seq}
     \includegraphics[width=0.23\textwidth]{pictures/conf_spred.png}} % \\[-1pt]
    \subfigure[]{
     \label{fig:conf_spred_NOTseq}
     \includegraphics[width=0.23\textwidth]{pictures/conf_spred_NOTseq.png}} % \\[-5pt]
\vspace{0pt}
   \caption[Confusion matrices for Stress-Predict using 10-fold CV (within-subject).] {\label{fig:conf_spred} Confusion matrices for  Stress-Predict using sequential \subref{fig:conf_spred_seq} and non-sequential  \subref{fig:conf_spred_NOTseq} 10-fold CV (within-subject), using contextual "stimuli" labels for seeding.}
   \end{figure}
\end{comment}

\begin{figure}[h!]
    \centering
    \includegraphics[width=0.48\textwidth]{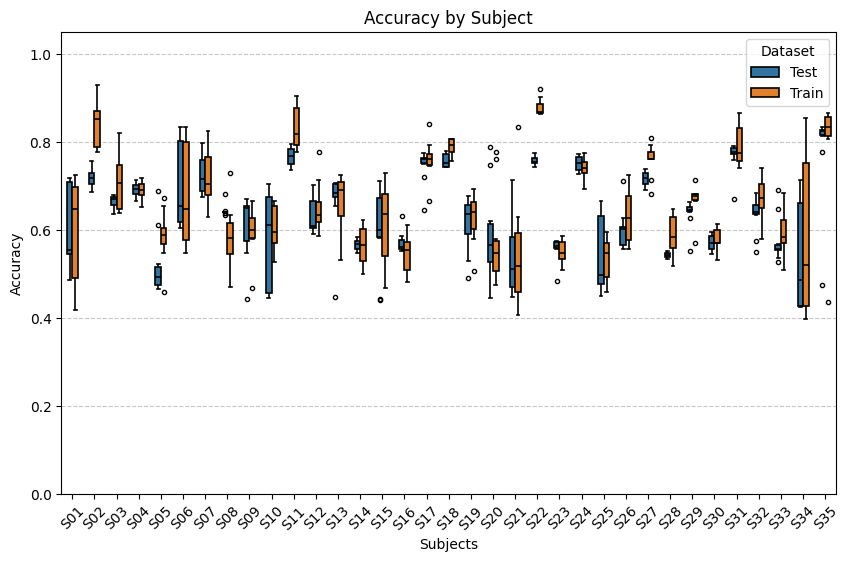}
    \caption{The non-sequential 10-fold CV accuracy for individual subjects of the Stress-Predict dataset in the test (blue) and train (orange) sets.}
            \label{fig:acc_spred_NOTseq}
\end{figure}

Figure \ref{fig:acc_spred_NOTseq} illustrates the accuracy results in Stress-Predict for individual subjects. It shows that, for most subjects, both the median accuracy and the 25\% quartile (across folds) are above 60\% at testing. For 12 out of the 35 subjects (S01, S05, S14, S16, S20, S21, S23, S25, S28, S30, S33, S34), both are below 60\%, showing poor model performance, while for 5 others (S09, S10, S15, S19, S26) the median value is above 60\% but the 25\% quartile is below this threshold.

The model performance results are similar at training and testing as the two ranges most often intersect and there is no significant difference between median results for training versus testing. Thus, the model shows no signs of overfitting. To understand whether the poor model performance results on some participants could be due to the fact that no hyperparameter tuning was performed, we did a sensitivity analysis testing different model parameters. These results are presented in section \ref{sec:senstivity}.

\begin{comment}
    \begin{figure}[h!]
    \centering
    \includegraphics[scale=0.5]{pictures/conf_spred.jpg}
    \caption{Confusion matrix for Stress-Predict using 10-fold CV (within-subject).}
\end{figure}

\textcolor{gray}{- seeding: only context (Stress vs. Baseline) - discuss results}
\end{comment}

\hfill

\subsubsection{\textbf{CEAP360-VR Dataset}}

Self-reported retrospective data was used for seeding in CEAP360-VR, and, after regrouping (due to size limitations, as only 8 videos are used, of around 1min duration each), two different classes were considered: stress and non-stress. This led to a binary classification problem (rather than multi-class), for which the model attained an accuracy of 54.6\% using sequential CV and 61.0\% using non-sequential CV, showing reasonable performance - while below the 67.1\% obtained in the original paper \cite{ceap_vr}.

\begin{figure}[h!]
    \centering
    \includegraphics[width=0.38\textwidth]{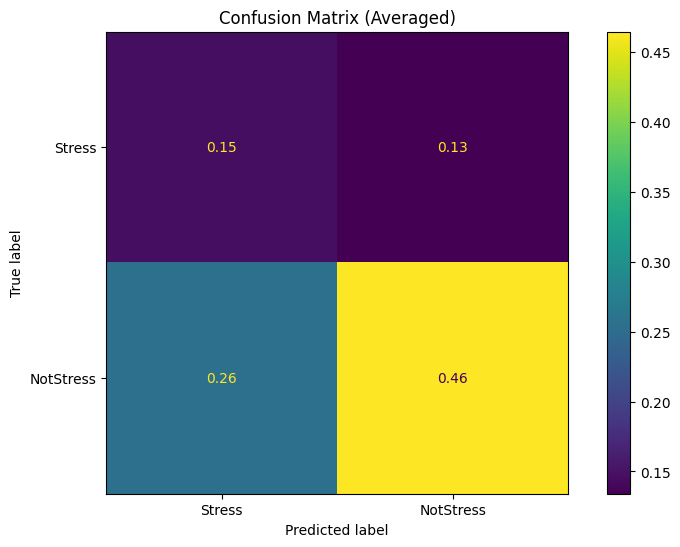}
    \caption{Confusion matrix for CEAP360-VR using non-sequential 10-fold CV (within-subject), considering self-reported "emotional" labels for seeding.}
        \label{fig:conf_ceapvr_NOTseq}
\end{figure}

Notably, only 29 out of the 32 subjects in the dataset were considered, as the remaining 3 did not assign any of the videos to the stress quadrant. The confusion matrix (averaged across subjects) is shown in Figure \ref{fig:conf_ceapvr_NOTseq}. 

While ensuring sufficient data points from all classes, the regrouping into 2 single categories resulted in class imbalance, with the majority of data points being pseudo-labeled as non-stress (both in the emotional and contextual seeding approaches). In some cases, this led the model to simply classify all data samples as the most frequent class (non-stress), achieving good accuracy results even though it could not learn to identify and distinguish between the different affective states elicited by the videos (e.g. for participant P32). This explains why, even though the overall model accuracy is reasonable, low precision and recall metrics were obtained, as previously summarized in Table \ref{table:model_res}.

\begin{figure}[h!]
    \centering
    \includegraphics[width=0.48\textwidth]{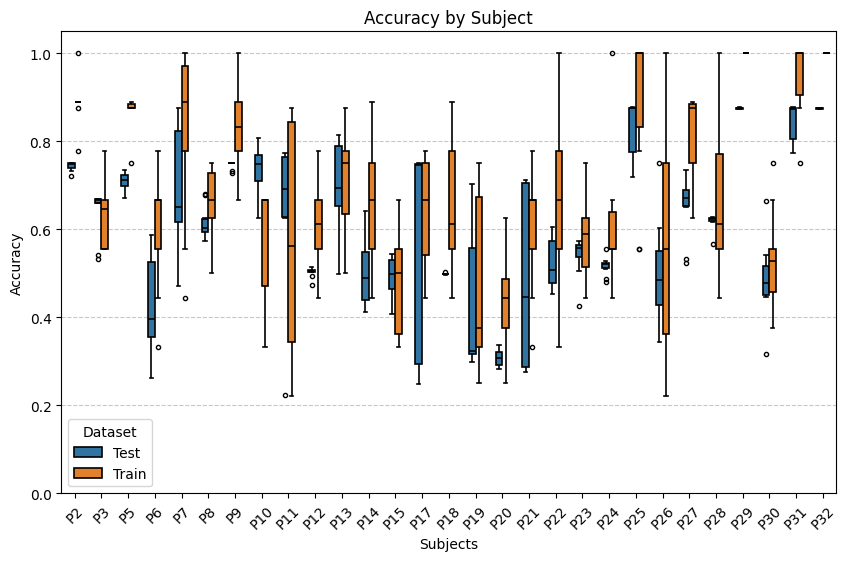}
    \caption{The non-sequential 10-fold CV accuracy for individual subjects of the CEAP360-VR dataset in the test (blue) and train (orange) sets.}
    \label{fig:acc_ceap_NOTseq}
\end{figure}

The accuracy results for all individual subjects are presented in Figure \ref{fig:acc_ceap_NOTseq}. The contextual "stimuli" seeding approach was also tested, but led to worse model performance (both globally and considering only this subset of 29 participants).

\begin{comment}
\begin{figure}[h!]
   \vspace{0pt}
%\smallskip
   %\centering
    \subfigure[]{
     \label{fig:conf_ceapvr}
     \includegraphics[width=0.23\textwidth]{pictures/conf_ceapvr.png}} % \\[-1pt]
    \subfigure[]{
     \label{fig:conf_ceapvr_NOTseq}
     \includegraphics[width=0.23\textwidth]{pictures/conf_ceapvr_NOTseq.png}} % \\[-5pt]
\vspace{0pt}
   \caption[Confusion matrices for CEAP360-VR using 10-fold CV (within-subject).] {\label{fig:conf_ceapvr} Confusion matrices for CEAP360-VR using sequential \subref{fig:conf_ceapvr} and non-sequential  \subref{fig:conf_ceapvr_NOTseq} 10-fold CV (within-subject), using self-reported "emotional" labels for seeding.}
   \end{figure}
\end{comment}

\begin{comment}
\begin{figure}[h!]
    \centering
    \includegraphics[scale=0.4]{pictures/acc_ceapvr.png}
    \caption{The sequential 10-fold CV accuracy for individual subjects of the CEAP360-VR dataset.}
\end{figure}
\end{comment}

\begin{comment}

CONTEXT

53.3\% with context - sample /2000 in training - 31 subjects

\begin{figure}[h!]
   \vspace{-12pt}
%\smallskip
   %\centering
    \subfigure[]{
     \label{fig:conf_ceapvr_context}
     \includegraphics[width=0.23\textwidth]{pictures/conf_ceapvr_context.png}} % \\[-1pt]
    \subfigure[]{
     \label{fig:conf_ceapvr_NOTseq_context}
     \includegraphics[width=0.23\textwidth]{pictures/conf_ceapvr_NOTseq_context.png}} % \\[-5pt]
\vspace{-5pt}
   \caption[Confusion matrices for CEAP360-VR using 10-fold CV (within-subject).] {\label{fig:conf_ceapvr_context} Confusion matrices for CEAP360-VR using sequential \subref{fig:conf_ceapvr_context} and non-sequential  \subref{fig:conf_ceapvr_NOTseq_context} 10-fold CV (within-subject), using contextual "stimuli" labels for seeding.}
   \end{figure}

\begin{figure}[h!]
    \centering
    \includegraphics[scale=0.5]{pictures/conf_3rdataset.jpg}
    \caption{Confusion matrix for CEAP360-VR using 10-fold CV (within-subject).}
\end{figure}
\end{comment}

\subsection{Sensitvity Analysis} \label{sec:senstivity}

As mentioned in section \ref{sec:res_spred}, a sensitivity analysis was made considering the Stress-Predict dataset, in order to assess whether poor model performance results in some participants could be due to the use of non-optimal model parameters. Different values for sequence length (previously 600) and embedding size (previously 30) were tested.

\begin{figure}[h]
    \centering
    \subfigure[]{
    \includegraphics[width=0.48\textwidth]{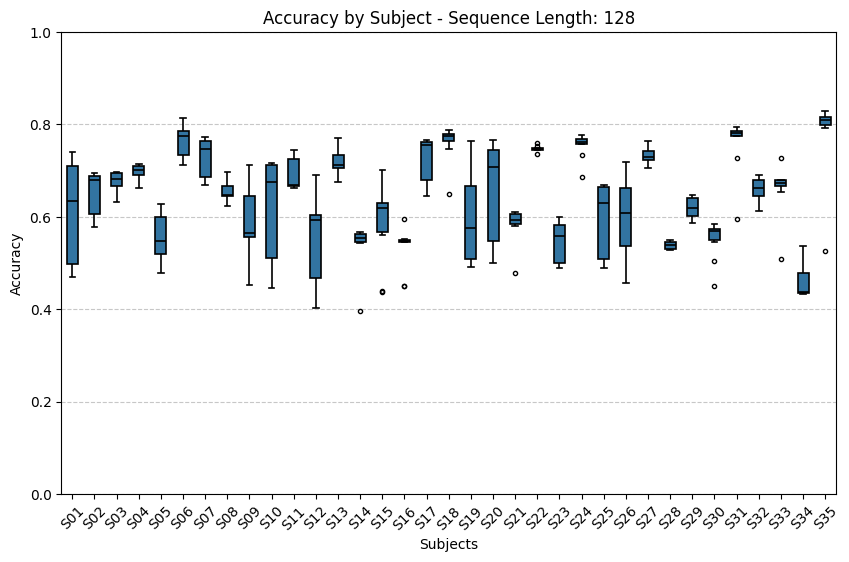}
    \label{128}}
        \centering
    \subfigure[]{
    \includegraphics[width=0.48\textwidth]{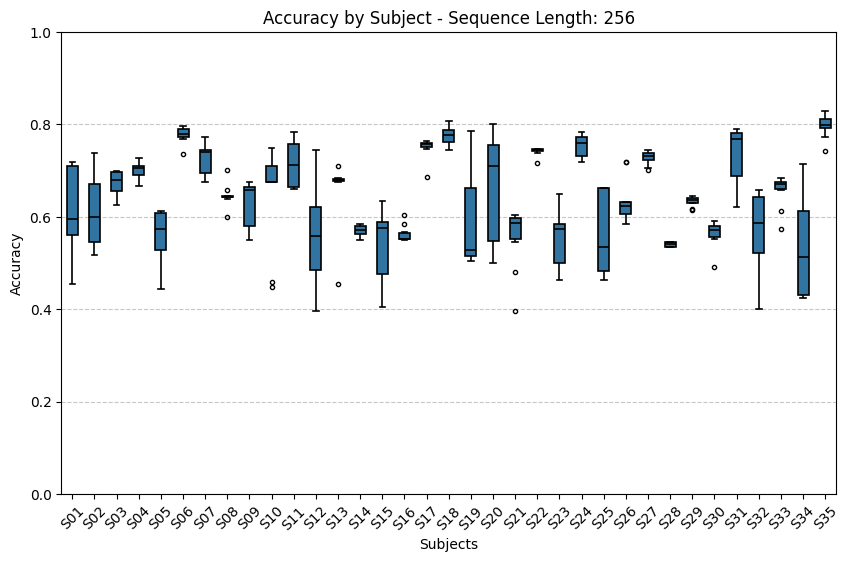}
        \label{256}}
        \centering
    \subfigure[]{
    \includegraphics[width=0.48\textwidth]{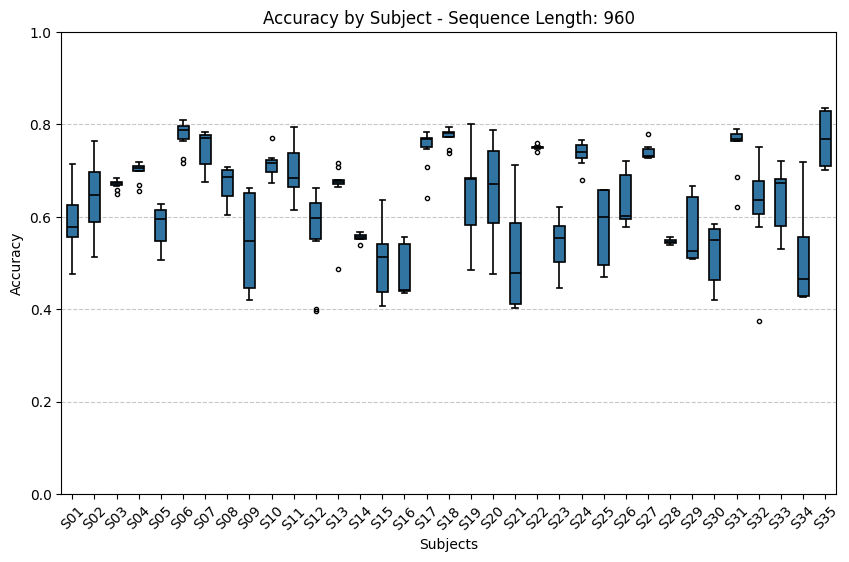}
            \label{960}}
    \caption{The non-sequential 10-fold CV accuracy in the test set, for individual subjects of the Stress-Predict dataset, considering different model parameters (sequence length, in samples: \ref{128} 128,   \ref{256}, 256, and \ref{960} 960).}
            \label{fig:acc_spred_sensitivity_seqlen}
\end{figure}

\begin{figure}[h]
    \centering
    \subfigure[]{
    \includegraphics[width=0.48\textwidth]{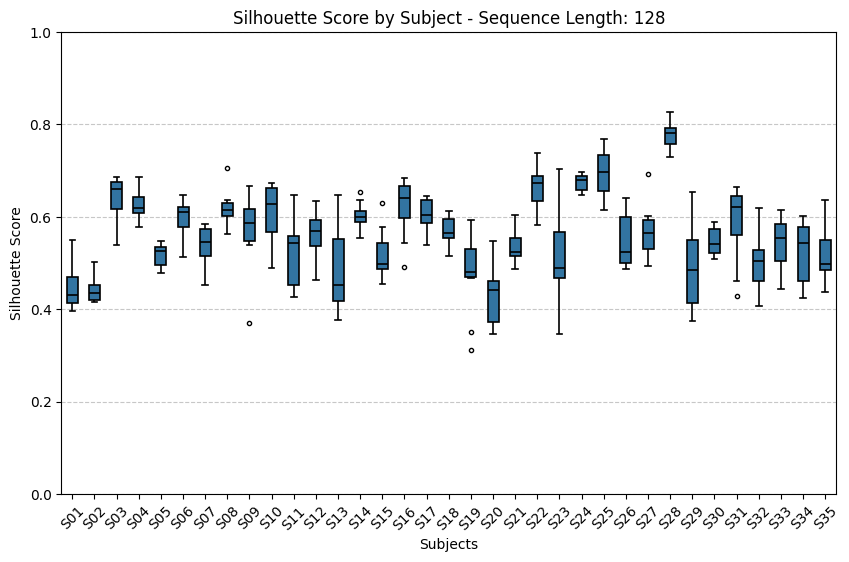}\label{silh128}}
        \centering
    \subfigure[]{
    \includegraphics[width=0.48\textwidth]{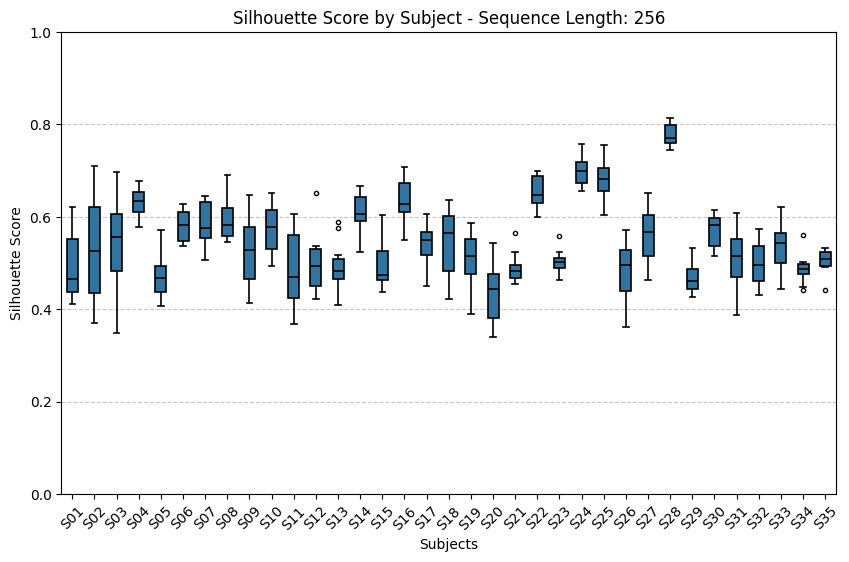}\label{silh256}}
        \centering
    \subfigure[]{
    \includegraphics[width=0.48\textwidth]{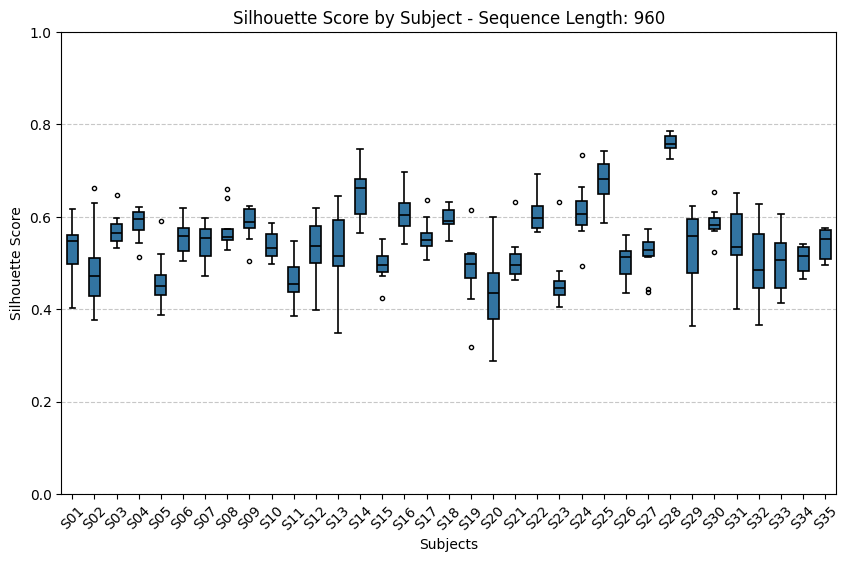}\label{silh960}}
    \caption{The non-sequential 10-fold CV silhouette scores, for individual subjects of the Stress-Predict dataset, considering different model parameters (sequence length, in samples: \ref{silh128} 128, \ref{silh256}, 256, and \ref{silh960} 960).}
            \label{fig:silh_spred_sensitivity_seqlen}
\end{figure}

In line with the research from \cite{ekman2007recognizing}, who concluded that the duration of emotion typically ranges from 0.5s to 4s, we tested shorter sequence lengths (128 and 256, equivalent to 1s and 2s), and also a longer duration of 15s (960 samples) which should reflect mood, rather than emotion. The results considering those different sequence lengths are shown in Figures \ref{fig:acc_spred_sensitivity_seqlen} (accuracy) and  \ref{fig:silh_spred_sensitivity_seqlen} (silhouette scores).

While the average classification results across subjects for each fixed parameter remain close to 64.2\%, this change is more significant if, instead, the optimal model parameters (out of the 4 tested sequence lengths) are selected for each particular subject - in line with the fact that the model was trained using a within-subject approach. This results in an average accuracy of 66.7\%, proving that model performance is largely dependent on the selected parameters. The individual subject results for the four different sequence lengths are presented in Table \ref{table:acc_seqlen}.

\begin{table}[h]
\begin{center}
\caption{Average test accuracy (\%) using different sequence lengths}
\begin{tabular}{ |c|c|c|c|c|}
\hline
        & seq 128 & seq 256  & seq 600 & seq 960      \\
\hline
S01     & 61.4        & \textbf{61.8}       & 60.8 & 59.3    \\
S02     & 65.2        & 61.1       & \textbf{71.9} & 63.6    \\
S03     & \textbf{67.8}        & 67.5       & 66.6 & 67.1    \\
S04     & 69.8        & \textbf{70.1}       & 69.2 & 69.8    \\
S05     & 55.6        & 56.2       & 52.0 & \textbf{57.9}    \\
S06     & 76.5        & \textbf{77.8}       & 70.0 & 77.6    \\
S07     & 72.9        & 72.4       & 72.5 & \textbf{74.7}    \\
S08     & 65.7        & 64.6       & 64.4 & \textbf{66.9}    \\
S09     & 59.1        & \textbf{63.0}       & 61.3 & 54.7    \\
S10     & 62.4        & 64.9       & 57.9 & \textbf{71.2}    \\
S11     & 69.3        & 71.3       & \textbf{76.6} & 69.9    \\
S12     & 56.0        & 55.9       & \textbf{63.2} & 56.6    \\
S13     & \textbf{72.0}        & 66.0       & 66.4 & 66.3    \\
S14     & 54.0        & \textbf{57.1}       & 56.6 & 55.6    \\
S15     & 58.6        & 53.7       & \textbf{60.0} & 50.0    \\
S16     & 53.4        & 56.1       & \textbf{57.2} & 48.1    \\
S17     & 72.2        & \textbf{75.0}       & 74.7 & 75.0    \\
S18     & 76.1        & \textbf{77.6}       & 75.8 & 77.5    \\
S19     & 60.1        & 58.8       & 61.7 & \textbf{64.5}    \\
S20     & 66.0        & \textbf{66.5}       & 58.8 & 65.3    \\
S21     & \textbf{58.5}        & 55.7       & 53.4 & 51.6    \\
S22     & 74.7        & 74.2       & \textbf{75.7} & 75.0    \\
S23     & 54.7        & \textbf{55.9}       & 55.8 & 54.2    \\
S24     & \textbf{75.5}        & 75.3       & 75.1 & 73.7    \\
S25     & \textbf{59.5}        & 56.2       & 54.1 & 57.9    \\
S26     & 60.6        & \textbf{63.4}       & 60.1 & 63.2    \\
S27     & 73.2        & 72.8       & 71.5 & \textbf{74.0}    \\
S28     & 53.9        & 54.0       & 54.4 & \textbf{54.7}    \\
S29     & 61.9        & 63.4       & \textbf{63.8} & 57.0    \\
S30     & 55.2        & 56.4       & \textbf{57.1} & 52.3    \\
S31     & 76.0        & 73.2       & \textbf{76.8} & 75.0    \\
S32     & \textbf{66.0}        & 56.9       & 63.4 & 63.0    \\
S33     & \textbf{66.1}        & 65.6       & 57.6 & 64.1    \\
S34     & 46.0        & \textbf{53.6}       & \textbf{53.6} & 51.5    \\
S35     & 78.2        & \textbf{79.8}       & 78.5 & 76.9    \\
\hline
\end{tabular}
\label{table:acc_seqlen}
\end{center}
\end{table}

\begin{figure}[h]
    \centering
    \subfigure[]{
    \includegraphics[width=0.48\textwidth]{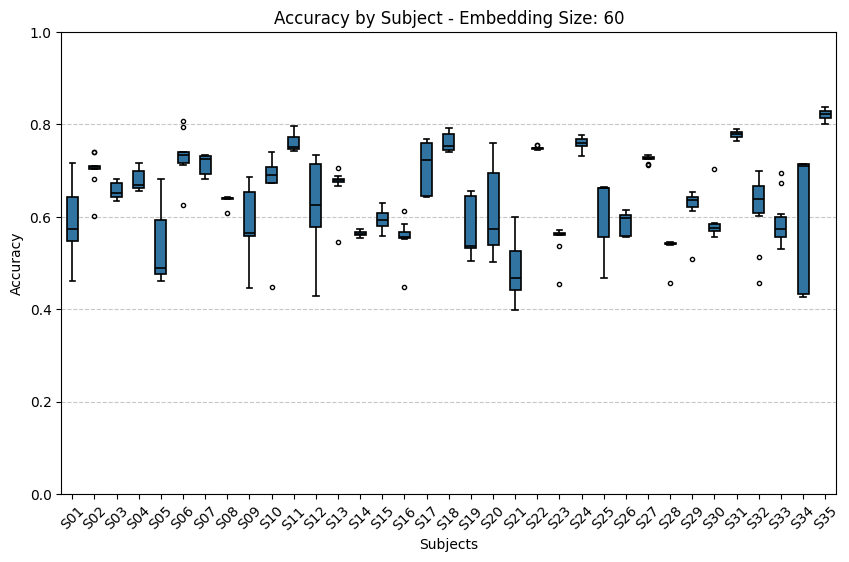}
    \label{60}}
        \centering
    \subfigure[]{
    \includegraphics[width=0.48\textwidth]{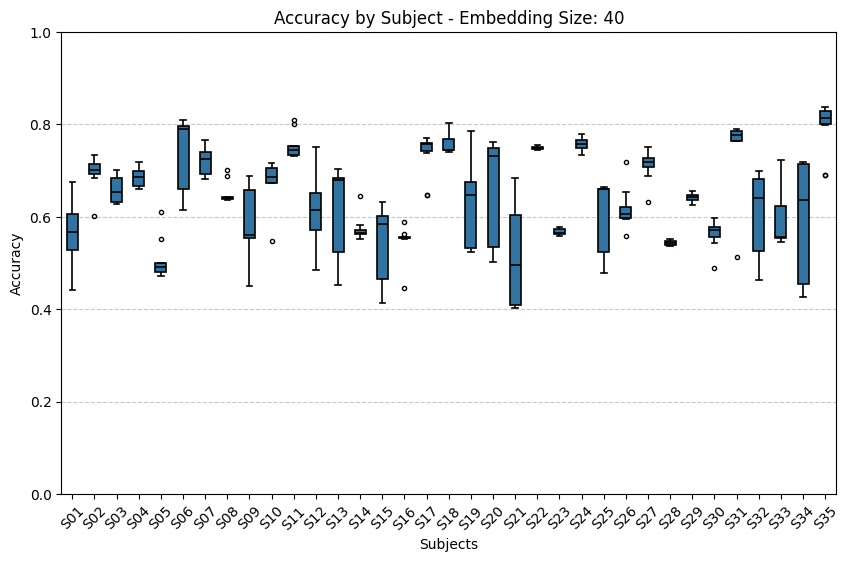}
        \label{40}}
        \centering
    \caption{The non-sequential 10-fold CV accuracy in the test set, for individual subjects of the Stress-Predict dataset, considering different model parameters (embedding sizes: \ref{60} 60 and \ref{40} 40).}
            \label{fig:acc_spred_sensitivity_emb}
\end{figure}

\begin{figure}[h]
    \centering
    \subfigure[]{
    \includegraphics[width=0.48\textwidth]{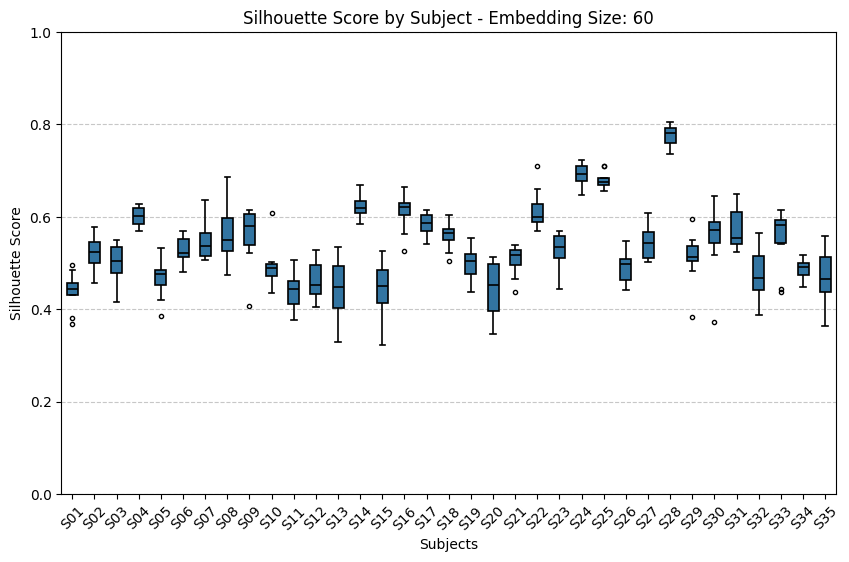}
    \label{silh60}}
        \centering
    \subfigure[]{
    \includegraphics[width=0.48\textwidth]{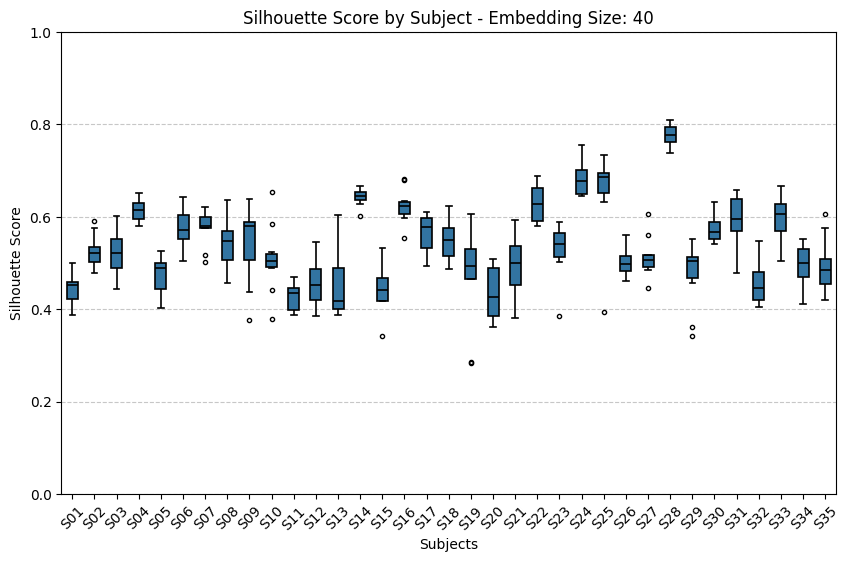}
        \label{silh40}}
        \centering
    \caption{The non-sequential 10-fold CV silhouette scores, for individual subjects of the Stress-Predict dataset, considering different model parameters (embedding sizes: \ref{silh60} 60 and \ref{silh40} 40).}
            \label{fig:silh_spred_sensitivity_emb}
\end{figure}

The results considering different embedding sizes, while keeping the sequence length fixed at the original 600 samples, are shown in Figures \ref{fig:acc_spred_sensitivity_emb} (accuracy) and  \ref{fig:silh_spred_sensitivity_emb} (silhouette scores). Once again, while the average classification results across subjects for each fixed parameter remain close to 64.2\%, it increases up to 65.7\% when selecting the optimal model parameters (out of the 3 tested embedding sizes) for each particular subject. The individual subject results for the four different sequence lengths are presented in Table \ref{table:acc_emb}.

\begin{table}[h]
\begin{center}
\caption{Average test accuracy (\%) using different embedding sizes}
\begin{tabular}{ |c|c|c|c|}
\hline
        & emb 60 & emb 40  & emb 30   \\
\hline
S01     & 59.0        & 57.2       & \textbf{60.8}     \\
S02     & 70.1        & 69.6       & \textbf{71.9}    \\
S03     & 65.7        & 66.0       & \textbf{66.6}     \\
S04     & 67.9        & 68.6       & \textbf{69.2}   \\
S05     & \textbf{53.1}        & 50.6       & 52.0    \\
S06     & 73.3        & \textbf{74.0}       & 70.0  \\
S07     & 71.4        & 72.2       & \textbf{72.5}   \\
S08     & 63.7        & \textbf{65.2}       & 64.4    \\
S09     & 59.0        & 59.0       & \textbf{61.3}  \\
S10     & 67.2        & \textbf{67.7}       & 57.9   \\
S11     & 76.1        & 75.4       & \textbf{76.6} \\
S12     & 62.6        & 61.5       & \textbf{63.2}    \\
S13     & \textbf{66.8}        & 61.8       & 66.4   \\
S14     & 56.4        & \textbf{57.3}       & 56.6     \\
S15     & 59.4        & 54.5       & \textbf{60.0}     \\
S16     & 55.5        & 54.8       & \textbf{57.2}    \\
S17     & 70.8        & 73.6       & \textbf{74.7}     \\
S18     & \textbf{76.1}        & 75.7      & 75.8   \\
S19     & 57.6        & \textbf{63.1}       & 61.7     \\
S20     & 61.1        & \textbf{65.8}       & 58.8  \\
S21     & 48.1        & 51.2       & \textbf{53.4}   \\
S22     & 74.9        & 75.0       & \textbf{75.7}    \\
S23     & 55.1        & \textbf{56.8}       & 55.8  \\
S24     & \textbf{75.9}        & 75.6       & 75.1   \\
S25     & \textbf{61.2}        & 60.4       & 54.1   \\
S26     & 58.6        & \textbf{61.7}       & 60.1  \\
S27     & \textbf{72.6}        & 71.1       & 71.5    \\
S28     & 53.4        & \textbf{54.4}       & \textbf{54.4}    \\
S29     & 62.3        & \textbf{64.3}       & 63.8   \\
S30     & \textbf{58.7}        & 56.3       & 57.1    \\
S31     & \textbf{77.8}        & 75.1       & 76.8    \\
S32     & 61.9        & 60.2       & \textbf{63.4}   \\
S33     & 58.9        & \textbf{59.0}       & 57.6  \\
S34     & \textbf{60.0}        & 59.4       & 53.6   \\
S35     & \textbf{82.1}        & 79.5       & 78.5   \\
\hline
\end{tabular}
\label{table:acc_emb}
\end{center}
\end{table}

We conclude that model performance could be optimized if considering hyperparameter tuning. Thus, future work on this or other datasets alike should accommodate this step.

\section{Conclusion}

In this work, we proposed and tested an end-to-end deep learning framework for emotion recognition that requires minimal supervision (i.e., labels) and no domain knowledge. We obtained reasonable to good model performance on three different validated datasets from the literature. A sensitivity analysis also showed that model performance could be improved using different parameters (e.g. sequence length and embedding size), so that hyperparameter tuning should be considered in future applications of the model.

One other limitation of the model comes from the fact that some assumptions are required for data clustering with k-means, and consequently, c-means. Particularly, k-means assumes that each cluster is approximately isotropic and well represented by a prototype (the centroid); that each data point is closer to the nearest point (to which the distance should ideally be small) in the same cluster than to all the ones in the remaining clusters; and that each of them has a uniform density of data. If these assumptions are not met, the c-means algorithm might not be able to capture and identify high quality clusters, reflecting in poor silhouette scores, and ultimately jeopardizing the performance of the full model. 

Despite these limitations, the model has great potential for application in outdoor and real-life settings, filling a gap on such naturalistic experiments. Indeed, it has proven to perform well on physiological data collected from non-intrusive wearable devices, showing good accuracy results not only in static video-watching contexts, but also in VR paradigms which include movement. As it also requires minimal supervision, it should be appropriate for longitudinal experiments where the number of prompts to the participants should be minimized at the risk of dropout. Thus, future work should be able to apply this model to outdoor, naturalistic, real-life settings.

\bibliographystyle{apalike}
\bibliography{bibliography.bib}

\end{document}